\documentclass[journal]{IEEEtran}
\usepackage{times}  
\usepackage{helvet}  
\usepackage{courier}  
\usepackage{url}  
\usepackage{graphicx}  
\usepackage{epsfig}
\usepackage{subfigure}
\usepackage{amsmath}
\usepackage{amssymb}
\usepackage{enumerate}
\usepackage{multirow}
\usepackage{algorithm}
\usepackage{algpseudocode}
\usepackage{graphics}
\usepackage{cite}
\usepackage{amsthm}
\theoremstyle{plain}
\newtheorem{thm}{Theorem}[section]
\newtheorem{lem}[thm]{Lemma}
\newtheorem{prop}[thm]{Proposition}

\theoremstyle{definition}

\theoremstyle{remark}

\begin{document}
\title{Manifold Adversarial Learning}
\author{Shufei~Zhang,
        Kaizhu~Huang*,
        Jianke~Zhu,
        Yang~Liu
\thanks{Shufei Zhang is with the Department of Electrical and Electronic Engineering, Xi'an Jiaotong-Liverpool University, Suzhou, 215123, China (e-mail: Shufei.Zhang@xjtlu.edu.cn).}%
\thanks{Kaizhu Huang is with the Department of Electrical and Electronic Engineering, Xi'an Jiaotong-Liverpool University, Suzhou, 215123, China (e-mail: Kaizhu.Huang@xjtlu.edu.cn).}
\thanks{Jianke~Zhu is with Department of Computer Science, Zhejiang University,Zhejiang, P.R.China (jkzhu@zju.edu.cn)}%
\thanks{Yang Liu is with Alibaba Group, China(panjun.ly@alibaba-inc.com)}
\thanks{Corresponding Author: Kaizhu Huang}
}

\markboth{Journal of \LaTeX\ Class Files, xx-xx, xx~2019}%
{Shell \MakeLowercase{\textit{et al.}}: Bare Demo of IEEEtran.cls for IEEE Journals}

\maketitle

\begin{abstract}
  Recently proposed adversarial training methods show the robustness  to both adversarial and original examples and  achieve  state-of-the-art results in supervised and semi-supervised learning.  All the existing adversarial training  methods consider only how the worst perturbed examples (i.e., adversarial examples) could affect the model output. Despite their success, we argue that such setting may be in lack of generalization, since the output  space (or label space)  is apparently less informative.
  In this paper, we propose a novel method, called Manifold Adversarial Training (MAT).  MAT manages to build an adversarial framework based on how the worst perturbation could affect the distributional manifold rather than the output space. Particularly,  a latent data space with the Gaussian Mixture Model (GMM) will be first derived.  On one hand, MAT tries to perturb the input samples in the way that would rough the distributional manifold the worst. On the other hand,  the deep learning  model is trained trying to  promote in the latent space the manifold smoothness,  measured by the variation of Gaussian mixtures (given the local perturbation around the data point). Importantly, since the latent space is more informative than the output space, the proposed MAT can learn better a robust and compact data representation, leading to further performance improvement. The proposed MAT  is important in that it can be considered as a superset of one recently-proposed discriminative feature learning approach called center loss. We conducted a series of experiments in both supervised and  semi-supervised learning on three benchmark data sets, showing that the proposed MAT can achieve remarkable performance,  much better than those of the state-of-the-art adversarial approaches. We also present a series of visualization which could generate further understanding or explanation on adversarial examples.

\end{abstract}


%
\IEEEpeerreviewmaketitle

\section{Introduction}
\IEEEPARstart{A}{dversarial} examples refer to augmented data points generated by imperceptible perturbation of input samples.  Being difficult to distinguish from real examples, such adversarial examples could however change the prediction of many of the best learning models including the state-of-the-art deep learning models~\cite{szegedy2013intriguing}\cite{goodfellow2014explaining}\cite{nguyen2015deep}. To alleviate such problems, researchers have proposed adversarial training, able to certify both the robustness on adversarial examples and the generalization on original examples. Adversarial training could be used in both supervised  and semi-supervised training.  In supervised adversarial learning,  the data labels  are needed to derive the worst perturbation against loss function~\cite{sinha2017certifiable}\cite{lyu2015unified}; in semi-supervised adversarial learning, a virtual adversarial training (VAT) is used by smoothing the output distribution with penalizing the $KL$-divergence between outputs of adversarial and original examples. VAT achieves state-of the-art performance on both image and text classification~\cite{miyato2017virtual}\cite{miyato2015distributional}\cite{miyato2016adversarial}.

Previous adversarial training methods simply consider how to make the results of prediction worse (in the output space) without considering how the data are robustly represented in a latent space. In general, the latent space is much more informative than the output space. It is hence meaningful if we can design the adversarial learning in the latent space rather than the output space. In this work, we develop a novel model called Manifold Adversarial Training (MAT) in the latent space.  We engage an information based regularization, i.e., Maximum Mutual Information (MMI)~\cite{bahl1986maximum}\cite{wells1996multi} so as to define a distributional manifold in the latent space. We then apply the adversarial training to smooth such manifold by penalizing the $KL$-divergence between the distributions of latent features of the adversarial and original examples. The novel framework is trained in an adversarial way: the adversarial noise is generated to rough the distributional manifold, while the model is trained to smooth it to make the latent space more representative. It is similar to traditional Laplacian regularization methods with locality-preserving properties~\cite{belkin2003laplacian}\cite{belkin2006manifold}. However, our approach is based on information geometry and the information metric $KL$-divergence.

To our best knowledge, this is one novel work that learns adversarially both a robust and compact representation in the latent space. It also presents a unified framework in that a simplified MAT could derive a famous recently-proposed discriminative feature learning model~\cite{wen2016discriminative}\cite{Wan2018Rethinking}. Though developed in the framework of supervised learning, it is straightforward and much easier to be extended in semi-supervised learning. We  develop a feasible and efficient training algorithm capable of  obtaining remarkably better performance than the existing adversarial training approaches. In particular, we  implemented our proposed method on benchmark datasets MNIST,  CIFAR-10, and SVHN. Our method achieves in supervised and semi-supervise learning the state-of-the-art performance, much better than the best of the existing counterpart methods.

\section{Related Work}
 It has a long history to use the perturbed examples to regularize the output~\cite{srivastava2014dropout}. Bishop et al. proposed a method to add the Gaussian noise to input samples and showed that it is equivalent to adding the penalty term to original objective function~\cite{bishop1995training}. Dropout can also be treated as random perturbation to prevent from over fitting~\cite{srivastava2014dropout}\cite{gal2016dropout}. The Unified Gradient Regularization Family is proposed to find the worst perturbation to increase the objective function~\cite{lyu2015unified}. It approximates the non-convex problem with Taylor series and applies the Lagrange multiplier method to evaluate the worst perturbation. Another similar work proposed by Aman et al. is to perturb the underlying data distribution in a Wasserstein ball~\cite{sinha2017certifiable}. Virtual Adversarial training proposed by Takeru~\cite{miyato2017virtual}\cite{miyato2016adversarial} is perhaps most related to our work. It developed a method extending the adversarial training to semi-supervised task by promoting the local smoothness of the output distribution.

Another type of semi-supervised learning methods are based on generative models. Ladder network combined the deep network and auto encoder with connections between two networks at each layer and achieves encouraging results~\cite{rasmus2015semi}. Triple generative Adversarial Network is proposed to combine Generative Adversarial Network (GAN) with classifier. There are three players, generator, discriminator and classifier playing against with each other~\cite{chongxuan2017triple}. Some Bayesian methods employ variational methods with deep learning~\cite{kingma2014semi}.

This work is also related to some traditional manifold regularization. Laplacian Eigenmaps was developed to consider how to construct a representation for data lying on a low dimensional manifold embedded in a high-dimensional space~\cite{belkin2003laplacian}. A geometric framework was also  proposed to exploit the geometry of the marginal distribution using the information of both labeled and unlabeled examples~\cite{belkin2006manifold}. Both of these two methods tried to smooth the low dimensional manifold with preserving local neighborhood information.

\section{Main Method}
In this section, we first introduce two previous adversarial training methods, adversarial training with $l_p$ norm constraint and Virtual Adversarial Training (VAT) which are close related to our proposed approach. Then we describe our proposed Manifold Adversarial Training (MAT). Specifically, we first introduce how to regularize the latent space with Maximum Mutual Information (MMI) and represent the latent features with Gaussian Mixtures. We then define the notion of the low dimensional submanifold of latent space and the smoothness of statistic manifold. After that we describe the major framework and present a series of theory to derive the practical optimization algorithm. Finally, we also briefly conduct a computational analysis. We now give a set of notations. The input set is represented as $\{x_i\}^N_{i=1}$ and the corresponding output is $\{y_i\}^N_{i=1}$. Let $x_i \in \mathcal{R}^I$ and $y_i \in \mathcal{R}^O$ where, $I$ is the dimension of input space and $O$ denotes the output dimension. The model distribution is denoted by $p_m(y|x, \theta)$ with model parameters $\theta$. In this paper, labeled dataset $D_l=\{x^i_l, y^i_l\}_{i=1}^{N_l}$ and unlabeled dataset $D_{ul}=\{x^i_{ul}\}_{i=1}^{N_{ul}}$ are used to train the model $p_m(y|x, \theta)$.

\subsection{Adversarial Training}
Adversarial training is to train the model with both the natural examples and adversarial examples. Goodfellow et al. proposed the Fast Gradient Sign Method (FGSM) to generate the adversarial examples~\cite{goodfellow2014explaining} and Lyu et al proposed a method which extends FGSM to $l_p$ norm constraint~\cite{lyu2015unified}. Then we can formulate the optimization problem of adversarial training in the form of distribution:

\begin{small}
\begin{eqnarray}
\min_{\theta}\max_{\epsilon}D[q(y|x_l), p_m(y|x_l)]+ D[q(y|x_l), p_m(y|x_l+\epsilon)]
\nonumber\\
\quad s.t. \|\epsilon\|_p \leq \sigma\qquad\qquad\qquad\qquad\qquad\qquad\qquad\qquad\qquad\qquad
\label{eq:Final}
\end{eqnarray}
\end{small}
where $q(y|x_l)$ is the true distribution of output label given input and $p_m(.)$ denotes the model distribution. $\epsilon$ is a perturbation added on the input within a small range. $\sigma$ is a small value. None negative function $D[.]$ is used to measure the divergence between true distribution and the model distribution. For examples, KL-divergence, f-divergence can be used to measure the divergence. The objective of adversarial training is to fit the true distribution with model distribution both on natural examples and adversarial examples.

To implement the adversarial training, the inner maximization problem need to be firstly solved to obtain the worst perturbation. For simplifying optimization, the first Taylor expansion of $D(.)$ is used as the approximation for original objective function. With the $l_p$ constraint, the worst perturbation can be approximated as:
\begin{small}
\begin{eqnarray}
\epsilon = \sigma sign(\nabla \mathcal{L})(\frac{\lvert \nabla \mathcal{L} \rvert}{\|\nabla \mathcal{L}\|_{p^\ast}})^{\frac{1}{p-1}}
\end{eqnarray}
\end{small}
where $p^\ast$ is the dual of p, i.e, $\frac{1}{p^\ast}+\frac{1}{p}=1$ and $\nabla \mathcal{L}$ the first derivative of loss function with respect to input $x$. When $p=\infty$, this method can be degraded to Fast Gradient Sign Method (FGSM). And the worst perturbation becomes:

\begin{small}
\begin{eqnarray}
\epsilon = \sigma sign(\nabla \mathcal{L})
\end{eqnarray}
\end{small}

For finding the worst perturbation, it is easy to compute $\nabla \mathcal{L}$ with backpropagation. Moreover it does not need a large amount of computation since it just needs one more backward process. After computing the adversarial perturbation, the model distribution $p_m(y|x, \theta)$ is trained to approximate the true distribution $q(y|x)$ on both natural and adversarial examples. It has been proved to be able to achieve the better generalization performance than the traditional training methods of deep neural network.

\subsection{Virtual Adversarial Training}
Another work similar to our method is Virtual Adversarial Training (VAT). In most cases, the full labeled information is not provided and it is important to make use of the information of unlabeled data. Different from the traditional adversarial training methods, Virtual Adversarial Training method can utilize both labeled and unlabeled data. The optimization problem of VAT can be formulated as:

\begin{small}
\begin{eqnarray}
\min_{\theta}\max_{\epsilon}D[p_m (y|x), p_m(y|x+\epsilon, \theta)]
\nonumber\\
s.t.\quad \|\epsilon\|_2 \leq \sigma\qquad\qquad\qquad\qquad\qquad
\label{eq:Final}
\end{eqnarray}
\end{small}
where, $x$ is either labeled data or unlabeled data. Different from traditional adversarial training, Virtual Adversarial Training minimize the divergence between two model distributions instead of the divergence between true distribution and model distribution. The objective of VAT is to smooth the output distribution around $x$ through the method of adversarial training. Specifically, inner optimization problem is to find the worst perturbation $\epsilon$ within a small range to make the model distributions of $x$ and $x+\epsilon$ the most different, oppositely, the outer one is to try to make two model distributions the same. Therefore, VAT does not require the label information and can be applied on unsupervised and semi-supervised tasks.

\subsection{Manifold Adversarial Training}
Both the traditional adversarial training and Virtual Adversarial Training methods attempt to find the worst perturbation to make the outputs of models worst. However, these two methods do not consider how the data is represented in latent space. Therefore, we design a new adversarial method Manifold Adversarial Training (MAT) to learn a good geometric structure of the data distribution for improving the robustness of representation in latent space. In following subsections, we first introduce how to regularize the latent space with Maximum Mutual Information (MMI). We then define the notion of smoothness of distributional manifold. After that we describe the major framework and present a series of theory to derive the practical optimization algorithm. Finally, we also briefly conduct a computational analysis.

\subsection{Modeling the Latent Space}
In the probability theory, the mutual information is to measure the mutual dependence between two variables. In other words, it measures how much knowing one of these variables reduces uncertainty about the other. For learning discriminative features with more information of labels, MMI is proposed to increase the mutual dependence between the latent features and the class label which can be seen as an information regularization. We define the label set $\{c_j\}_{j=1}^M$ and the latent features $\{z_i\}_{i=1}^N\in\Re^D$, where $M$ and $N$ denote the number of classes and samples respectively. $D$ is the dimensionality of the latent space. The mutual information between the event $C=c_j$ and the event $Z=z_i$ with respect to $\theta$ is given by:
\begin{small}
\begin{eqnarray}
\begin{aligned}
I_{\theta}(C=c_j, Z=\mathbf{z}_{i}) &= log \frac{P(C=c_j, Z=\mathbf{z}_{i})}{P(C=c_j)P(Z=\mathbf{z}_{i})}
\\ &= log P_{\theta}(\mathbf{z_i}|c_j)-log\sum_{j=1}^{M} P_{\theta}(\mathbf{z_i}|c_j)P(c_j)
\label{eq:1}
\end{aligned}
\end{eqnarray}
\end{small}
where $C$ and $Z$ are two random variables representing class label and latent feature respectively.

In this work, we assume the conditional probability $P_{\theta}(z_i|c_j)$, the multivariate Gaussian distribution, as Eq.~(2). We also assume that  the latent features are generated by Gaussian mixtures. The number of components of Gaussian mixtures is given as the class number in this paper.
\begin{small}
\begin{eqnarray}
\begin{aligned}
u_j(\mathbf{z}_{i}) = \frac{1}{(2\pi)^{D/2}|\Sigma_j|^{1/2}} exp\{-\frac{1}{2}(\mathbf{z}_{i}-\mathbf{\mu_j})^{T}\Sigma_{j}^{-1}(\mathbf{z}_{i}-\mathbf{\mu_{j}})\}
\label{eq:2}
\end{aligned}
\end{eqnarray}
\end{small}

In Eq.~(2), $\mathbf{\mu_j}$ is the class center for class $c_j$ and $\Sigma_j$ is corresponding covariance matrix. $D$ represents the dimension of latent space. In this work, we assume the class prior $P(c_j)$ is a constant $1/M$ ($M$ is the class number). Then, the latent feature can be represented by Gaussian mixtures, $u(\mathbf{z_i}) = \frac{1}{M} \sum_{j=1}^{M}u_j(z_i) = \frac{1}{M} \sum_{j=1}^{M} \mathcal{N}(\mathbf{z_i}|\mu_{j}, \Sigma_{j})$.  Again,the number of the components is set to the class number $M$. Each component of Gaussian mixtures denotes the probability of the latent feature $\mathbf{z_i}$ assigned to class $c_j$. We reformulate the mutual information $I_{\theta}(c_j, \mathbf{z_i})$ as follows:
\begin{small}
\begin{eqnarray}
I_{\theta}(C=c_j, Z=\mathbf{z}_{i}) &= &log \mathcal{N}(\mathbf{z_i}|\mathbf{\mu_{j}}, \Sigma_{j})-log\sum_{j=1}^{M} u_j(\mathbf{z_i})\qquad\label{eq:3}
\\ &=& M log P(c_j|\mathbf{z}_{i})\nonumber
\end{eqnarray}
\end{small}

In this case, maximizing the mutual information is equivalent to maximizing the log posterior $logP(c_j|\mathbf{z_i})$. The last term of Eq.~(\ref{eq:3}) is the log marginal distribution which can be seen as the normalization term. Therefore, we can just maximize the first term $log \mathcal{N}(\mathbf{z_i}|\mathbf{\mu_{j}}, \Sigma_{j})$. It is equivalent to making the latent features more compact or discriminate with respect to class centers. Then we define the information regularization in this paper:
\begin{small}
\begin{eqnarray}
\begin{aligned}
I_{\theta} &= \sum_{i=1}^{N}log \mathcal{N}(\mathbf{z_i}|\mathbf{\mu_{z_i}}, \Sigma_{z_i})
\label{eq:4}
\end{aligned}
\end{eqnarray}
\end{small}

One of our objectives is to maximize the above information regularization term so as to model a good latent space. In this paper, though we assume GMM over the latent space, it presents a generalized version to a recently proposed famous center loss model~\cite{wen2016discriminative}.  This can be readily obtained in Proposition 0.1.
\begin{prop}
     The center loss~\cite{wen2016discriminative}  can be viewed as the special case of our proposed information regularization.
\end{prop} \label{define:optimal_end_configuration}
\begin{proof}
     We reformulate the information regularization as:
       \begin{small}
        \begin{eqnarray*}
        \begin{aligned}
        I_{\theta} = \sum_{i=1}^{N}log\frac{1}{(2\pi)^{D/2}|\Sigma_{z_i}|^{1/2}} exp\{-\frac{1}{2}(\mathbf{z}_{i}-\mathbf{\mu_{z_i}})^{T}\Sigma_{z_i}^{-1}(\mathbf{z}_{i}-\mathbf{\mu_{z_i}})\}
        \label{eq:5}
        \end{aligned}
        \end{eqnarray*}
\end{small}
    If  we make the covariance matrix as identity, we can have:
     \begin{small}
        \begin{eqnarray*}
        I_{\theta} &= &\frac{1}{\mathcal{C}}\sum_{i=1}^{N} log(exp\{-\frac{1}{2}(\mathbf{z}_{i}-\mathbf{\mu_{z_i}})^{T}(\mathbf{z}_{i}-\mathbf{\mu_{z_i}})\})
        \\ \nonumber&=&\frac{1}{\mathcal{C}}\sum_{i=1}^{N}(\mathbf{z}_{i}-\mathbf{\mu_{z_i}})^2
        \label{eq:6}
        \end{eqnarray*}
        \end{small}
         where $\mathcal{C}$ is a constant. It is clear that the above last term changes into the center loss as defined in~\cite{wen2016discriminative}.
         \end{proof}

\textbf{Remarks.}
Center loss~\cite{wen2016discriminative} is also proposed to penalize latent features to be closer to class centers. However, it implicitly assumes an identity covariance matrix as shown in Proposition 0.1. In comparison, we
propose a more generalized regularization.  With such regularization, we can represent the latent features with Gaussian mixtures which can be easily employed on information geometry and information metric $KL$-divergence, which we show in the next subsection. Note that, in this work, we just employ such regularization term in the last hidden layer.

\subsection{Defining Low Dimensional Statistical Manifold}
\begin{figure}[h!]
\label{fig1}
\centering
\includegraphics[width=.4\textwidth]{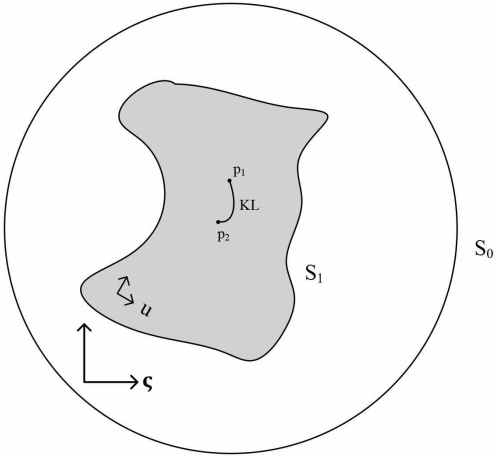}
\caption{The region of circle represents the entire manifold $S_0$ of latent space defined by the coordinate system $\zeta=(\zeta_1, \zeta_2, ... , \zeta_D)$ and the grey region in circle denotes a submanifold $S_1$ defined by the coordinate system $u=(u_1, u_2, ... , u_M)$. The degree of separation between two points $p_1$ and $p_2$ is described by $KL$-divergence.}\label{tsnetext1}
\end{figure}

An $n$-dimensional manifold $\mathcal{M}$ is defined as a set of points such that each point has $n$-dimensional extensions in its neighborhood and such a neighborhood is topologically equivalent to an $n$-dimensional Euclidean space. Intuitively speaking, it can be viewed as a deformed Euclidean space. In this paper, the dimension of the entire manifold $S_0$ of latent space is denoted as $D$ and the coordinate system is defined as  $\zeta=(\zeta_1, \zeta_2, ... , \zeta_D)$. In the previous section, the feature points are modeled by Gaussian mixtures. Hence, the set of all the Gaussian mixtures is an $M$-dimensional statistical submanifold $S_1$, where a point on manifold $S_1$ denotes a Gaussian mixture function and $u=(u_1, u_2, ... , u_M)$  describes a coordinate system ($u_i$ is the $i^{th}$ component of Gaussian mixture) as illustrated in Figure~\ref{tsnetext1}. The degree of separation of two points on manifold $S_1$ is measured by the $KL$-divergence between two Gaussian mixtures. In this paper, our aim is to smooth the manifold $S_1$.

\subsection{Defining Statistical Manifold Smoothing}
In this section, we consider  how to evaluate the smoothness of statistical manifold. First, we define some notations. The input set is denoted as $\{x_i\}_{i=1}^N$  and the output of model is defined as $\{y_i\}_{i=1}^N$. The training set is given as follows:
\begin{eqnarray}
\begin{aligned}
\mathcal{D} = \{(x_i, y_i)|x_i\in\Re^I, y_i\in\Re^O, n=1,...,N\}
\label{eq:7}
\end{aligned}
\end{eqnarray}

We then use $\mathcal{D}$ to train the model distribution $P_{\theta}(y|x,\theta)$ with the regularization term $I_{\theta}$. In the previous subsection, the latent features are represented by the Gaussian mixtures $u(z_i)$.  We can then easily define the distance between the two points on the statistical manifold with $KL$-divergence $KL[p||q]$ between mixtures $p$ and $q$. For learning a good manifold in the latent space, we try to preserve the local information in the embedding as in~\cite{belkin2003laplacian}\cite{belkin2006manifold}. Specifically, we add the small adversarial perturbation to input samples and try to make close the latent representation of perturbed samples and original one. Since the latent features are represented by Gaussian mixtures  in this work, the information metric $KL$-divergence can be readily  used to measure the similarity. The notion of smoothness of statistical manifold  can then be defined as the variation of the latent features on the statistical space caused by the adversarial perturbation $\epsilon$ as follows:
\begin{eqnarray}
\begin{aligned}
M_{KL}(x,\theta) \equiv KL[P_{GM}(f(x,\theta))||P_{GM}(f(x+\epsilon,\theta))]
\label{eq:8}
\end{aligned}
\end{eqnarray}
where  $f(x, \theta)$ denotes the latent feature of $x$ with the model parameters $\theta$. $P_{GM}(f(x))=\frac{1}{M} \sum_{j=1}^{M} \mathcal{N}(f(x)|\mu_{j}, \Sigma_{j})$ which is defined as the Gaussian mixture representation for latent features $f(x)$. $\epsilon$ represents the adversarial perturbation.

The smaller value of $M_{KL}(x,\theta)$ means the statistical manifold is more smooth at $x$ even when the perturbation  is imposed. We also define the smoothness in the output or label space as in~\cite{miyato2017virtual}\cite{miyato2016adversarial}:
\begin{eqnarray}
\begin{aligned}
L_{KL}(x,\theta) \equiv KL[P(y|x,\theta)||P(y|x+\epsilon,\theta)]
\label{eq:10}
\end{aligned}
\end{eqnarray}
In more details,  $L_{KL}(x,\theta)$ describes the $KL$-divergence  between the output space of adversarial example and the original examples. The overall smoothness for both the output and manifold  in latent space are then given as follows:
\begin{eqnarray}
\begin{aligned}
S_{KL}(x, \theta) \equiv L_{KL}(x, \theta) + M_{KL}(x, \theta)
\label{eq:11}
\end{aligned}
\end{eqnarray}

\subsection{Final Optimization Problem of MAT}
In the previous subsections,  we have defined the information regularization, manifold  smoothness,  and  label space smoothness.  We can now  obtain the final optimization problem of our proposed framework as follows:

\textbf{}
\begin{small}
\begin{eqnarray}
\max_{\theta}\min_{\epsilon}\frac{1}{N}\sum_{i=1}^{N} log P(y_i|x_i, \theta) - \lambda_1\frac{1}{N}\sum_{i=1}^{N} log S_{kl}(x_i, \theta) + \lambda_2I_{\theta}\nonumber\\
s.t.\quad \|\epsilon\|_2 \leq \sigma\qquad\qquad\qquad\qquad\qquad\qquad\qquad\qquad\qquad\qquad
\label{eq:Final}
\end{eqnarray}
\end{small}
where the hyperparameters $\lambda_1>0$, $\lambda_2>0$ and $\sigma>0$.

In the above objective function,  the first term represents the data log likelihood, the second term defines the overall data smoothness in both the manifold and output space, and the last term presents the information regularization. On one hand, the proposed MAT framework tries to optimize the model parameter $\theta$ so as to find the best latent space (by maximizing the information regularization $I_{\theta}$), enlarge  the data log likelihood  (so as to fit the data) as well as  increasing  the data smoothness (decreasing $\sum_{i=1}^{N} log S_{kl}(x_i, \theta)$); on the other hand, the imperceptible perturbation  $\epsilon$ tries to minimize the above objective function. In other words, the proposed novel model tries to find the best parameter $\theta$ that is even robust to the worst  perturbation as given by the optimal $\epsilon$. Since this adversarial learning is defined in the latent manifold space, we call this model as Manifold Adversarial Training (MAT).
In the next section, we will  use a strategy similar to~\cite{lyu2015unified} and~\cite{miyato2017virtual}  and  discuss how to solve this optimization problem practically and efficiently.

\subsection{Practical Algorithm}

Similar to~\cite{lyu2015unified}, we try to solve the above optimization problem in an alternative way. We first solve the inner minimization problem with respect to $\epsilon$, i.e. the worst perturbation, which we denote as $\epsilon_{adv}$. In order to calculate this worst perturbation, the $KL$-divergence between GMMs needs to be calculated firstly. Since it is difficult to conduct the evaluation directly, we approximate it by matching between the Gaussian elements of the two Gaussian mixture density as described in~\cite{goldberger2003efficient}:
\begin{eqnarray*}
\begin{aligned}
KL_{match}(f||g) &= \mathop{\min}_{\Psi}\sum_{i=1}^{K}\sum_{j=1}^{K}\alpha_{i}\Psi_{ij}(log\frac{\alpha_i}{\beta_j}+KL(f_i||g_j)) \\ &\leq\sum_{i=1}^{K}KL(f_i||g_i))
\label{eq:13}
\end{aligned}
\end{eqnarray*}
where $\Psi_{ij}$ is a $K\times K$ stochastic matrix. For simplifying the calculation, we try to optimize the upper bound of $KL$-divergence with assuming $\Psi_{ij}$ an identity matrix and the mixture weight $\alpha=1/K$ and $\beta=1/K$.

Since it is difficult to evaluate the worst perturbation in a non-convex problem, we  relax it to a convex problem with second-order Taylor expansions as in~\cite{miyato2017virtual}\cite{miyato2016adversarial}. Since $M_{KL}(\epsilon, x, \theta)$ reaches the minimum value when $\epsilon = 0$ and $P(y|x, \theta)$ is differentiable with respect to $x$ and $\theta$, the first derivative $\nabla_{\epsilon} M_{KL}(\epsilon, x, \theta)=0$. We can finally approximate it as follows:
\begin{eqnarray}
\begin{aligned}
\epsilon_{adv}(x, \theta) &\cong \mathop{\arg\max}_{\epsilon} \{\epsilon^T H(x,\theta) \epsilon; \|\epsilon\|_2\leq \sigma \} \\ &=\sigma \hat{d}
\label{eq:14}
\end{aligned}
\end{eqnarray}
where, $H(x, \theta)$ is the Hessian matrix calculated by employing the second derivative, $H(x, \theta)\equiv \nabla\nabla_{\epsilon}M_{KL}(\epsilon, x, \theta)|_{\epsilon=0}$.

Before we could obtain the worst perturbation, we first present  Lemma~\ref{lem1} and Lemma~\ref{lem2} as follows:
\begin{lem}
     Let $B$ be a real square matrix in $\mathbb{R}\times \mathbb{R}$ and $u$ be the dominant eigenvector of $B$ and $e$ is a vector which is not perpendicular to $u$. Then, the iterative calculation of

     \begin{eqnarray}
     \begin{aligned}
       e\gets \overline{Be}
     \label{eq:20}
     \end{aligned}
     \end{eqnarray}
     will make $e$ converge to $u$. ( $\overline{.}$ represents the normalization operator)
     \label{lem1}
 \end{lem}

\begin{lem}
     Let $A$ be a Hessian matrix function $F(x)$ with respect to $x$ in $\mathbb{R}\times \mathbb{R}$ and $r$ be a vector in $\mathbb{R}$. Then we have:

     \begin{eqnarray}
     \begin{aligned}
      Hr= \lim_{a\to 0} \frac{\nabla_x F(x+ar) - \nabla_x F(x)}{a}
     \label{eq:21}
     \end{aligned}
     \end{eqnarray}

\begin{proof}
We give the Taylor expansion of the first derivative of function $F(x+ar)$:

\begin{eqnarray}
\begin{aligned}
\nabla_x F(x+ar) = \nabla_x F(x) + Har + o(\|ar\|_2)
\label{eq:22}
\end{aligned}
\end{eqnarray}
then we have:

\begin{eqnarray}
\begin{aligned}
Hr &= \frac{\nabla_x F(x+ar) - \nabla_x F(x)}{a} + o(\|ar\|_2)\\
&= \lim_{a\to 0} \frac{\nabla_x F(x+ar) - \nabla_x F(x)}{a}
\label{eq:23}
\end{aligned}
\end{eqnarray}

\end{proof}

     \label{lem2}
 \end{lem}

By using Lemma 0.2 and Lemma 0.3 , we could solve the inner minimization problem, i.e., obtain the current worst perturbation. Specifically,  we have Theorem 0.4 showing that it  can then be written as the production of most sensitive direction $\hat{d}$ and the scale $\epsilon$, where the most sensitive direction $\hat{d}$ can be approximated iteratively by the power method.
\begin{eqnarray}
\begin{aligned}
d\gets\overline{\nabla_{\epsilon}M_{KL}(\epsilon, x, \theta)|_{\epsilon=\xi d}}
\label{eq:15}
\end{aligned}
\end{eqnarray}
where, $\overline{.}$ represents the normalization operator and $\xi>0$. $d$ is a vector which is not perpendicular to dominant eigenvector of $H(x,\theta)$.
\begin{thm}
     The steepest direction $\hat{d}$ in~(\ref{eq:14}) can be approximated by iterative calculation of:

      \begin{eqnarray}
      \begin{aligned}
      d\gets\overline{\nabla_{\epsilon}M_{KL}(\epsilon, x, \theta)|_{\epsilon=\xi d}}
      \label{eq:16}
      \end{aligned}
      \end{eqnarray}
     \label{thm1}
     where $\overline{.}$ represents the normalization operator.
     \begin{proof}
     Applying the Lagrange multiplier method on~(\ref{eq:14}), we can easily get that the steepest direction $\hat{d}$ has the same direction with the dominant eigenvector of $H(x,\theta)$. Then we can easily approximate the steepest direction $\hat{d}$ by iteritive calculation of (refer to Lemma 0.2):

     \begin{eqnarray}
     \begin{aligned}
     d\gets \overline{Hd}
     \label{eq:17}
     \end{aligned}
     \end{eqnarray}
     where, $d$ is a vector which is not perpendicular to dominant eigenvector of $H(x,\theta)$. Then we can calculate $Bv$ using Lemma 0.3:

    \begin{eqnarray}
     \begin{aligned}
      Hd&= \lim_{a\to 0} \frac{\nabla_{\epsilon}M_{KL}(\epsilon, x, \theta)|_{\epsilon=a d} - \nabla_{\epsilon}M_{KL}(\epsilon, x, \theta)|_{\epsilon=0}}{a} \\&\cong \nabla_{\epsilon}M_{KL}(\epsilon, x, \theta)|_{\epsilon=\xi d}
     \label{eq:18}
     \end{aligned}
     \end{eqnarray}
    where, $\xi$ is a small value. Then we have:

    \begin{eqnarray}
      \begin{aligned}
      d\gets\overline{\nabla_{\epsilon}M_{KL}(\epsilon, x, \theta)|_{\epsilon=\xi d}}
      \label{eq:19}
      \end{aligned}
      \end{eqnarray}

     \end{proof}

 \end{thm}

 After we solve the inner minimization problem with respect to $\epsilon$, we can then solve the maximization problem with respect to $\theta$. We iterate these two steps until the process converges. The detailed  pseudo code algorithm is shown in~Algorithm~\ref{alg:Framwork}.
\begin{figure*}[t]
\label{fig2}
\centering
\subfigure[The distribution of latent features from center loss] {
\psfig{file=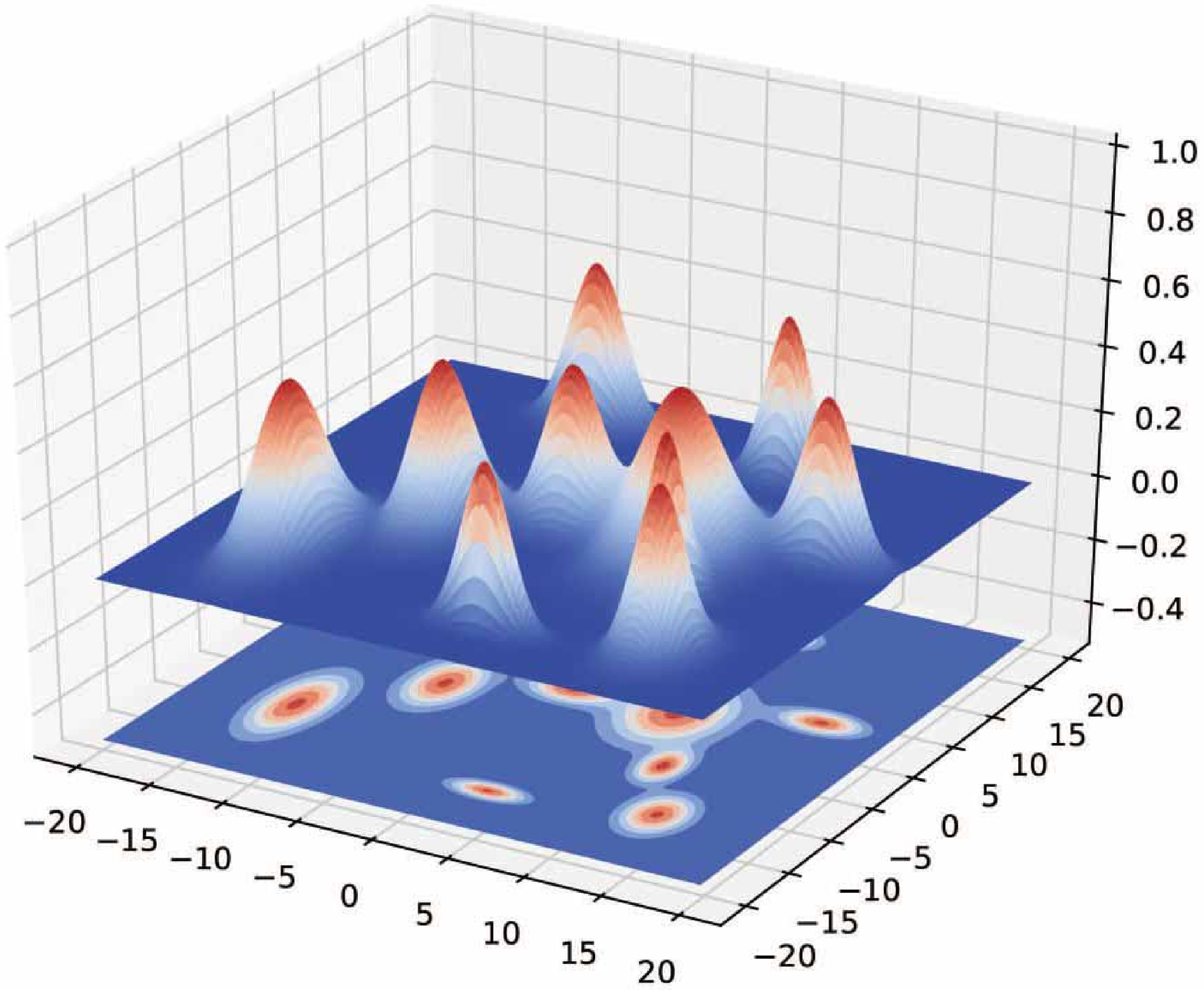,angle=0,width=0.225\textwidth}
\label{fig:tsne6} }
\subfigure[The distribution of latent features from MAT] {
\psfig{file=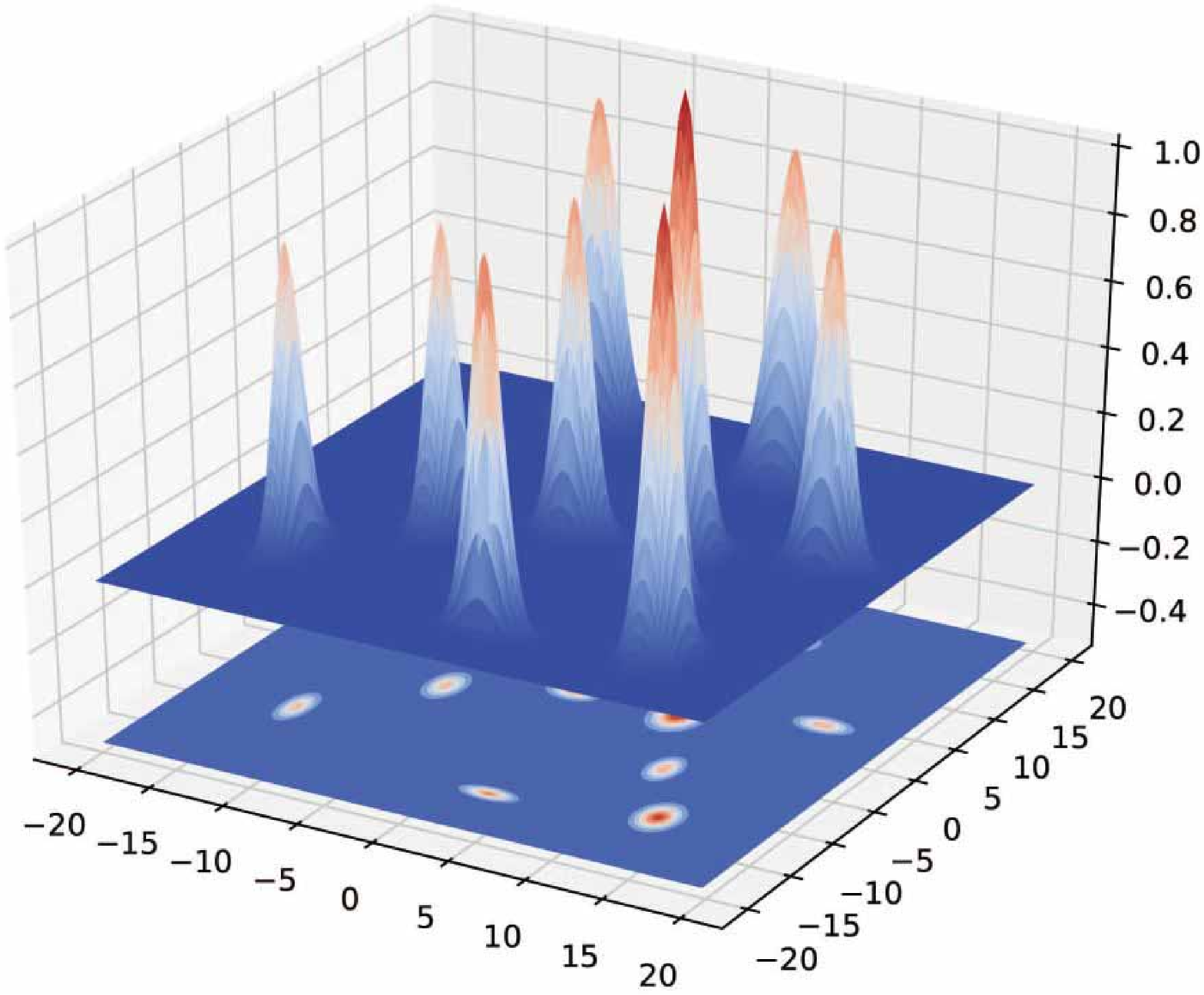,angle=0,width=0.225\textwidth}
\label{fig:tsne8} }
\subfigure[The distribution of latent features from softmax] {
\psfig{file=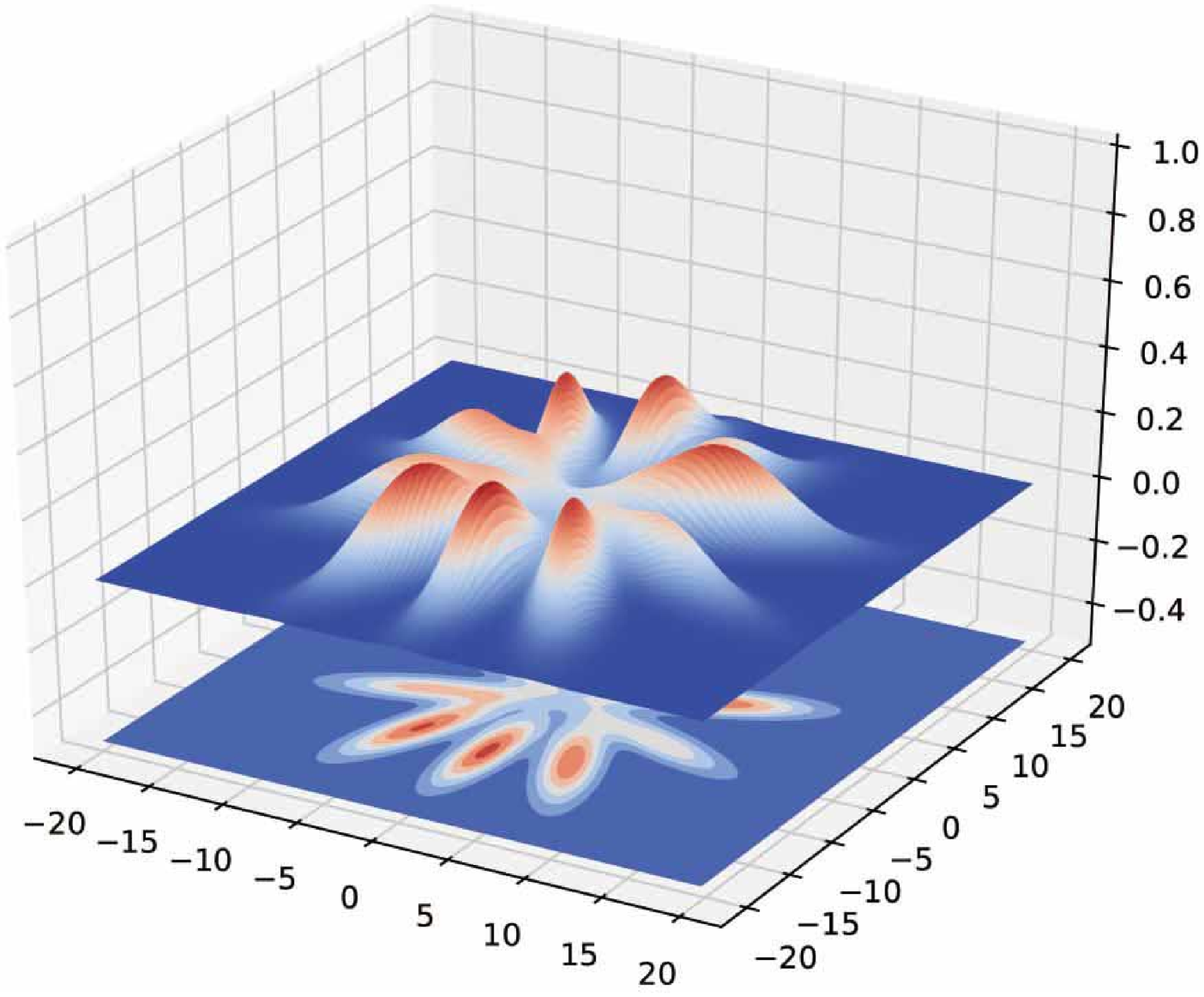,angle=0,width=0.225\textwidth}
\label{fig:tsne9} }
\subfigure[The distribution of latent features from VAT] {
\psfig{file=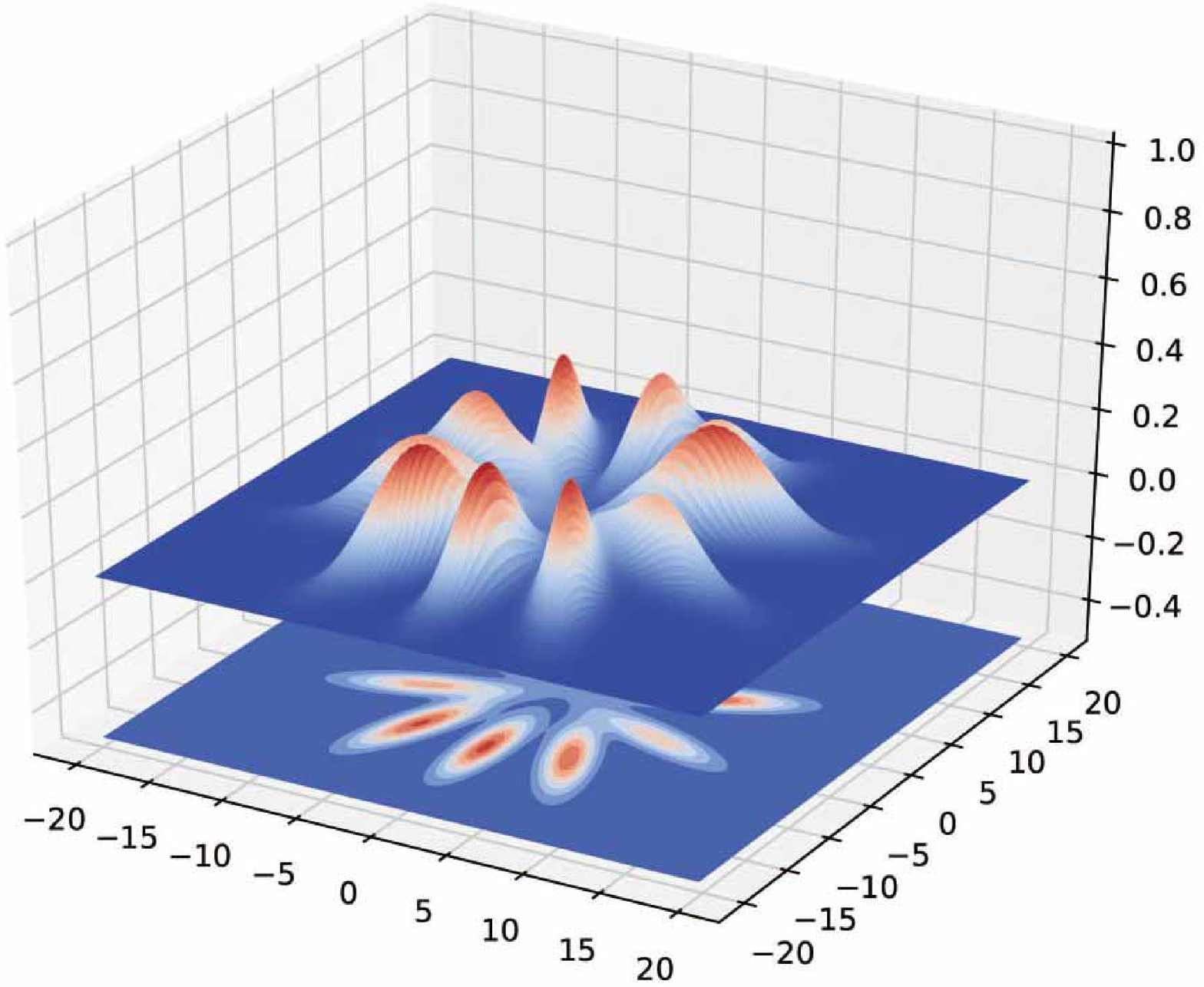,angle=0,width=0.225\textwidth}
\label{fig:tsne10} }
\caption{Embedding space of Lenet++ of different methods on MNIST test data. The graphs are better viewed in color.}\label{tsnetext}
\end{figure*}

\begin{algorithm}[h!]
\centering
\caption{Algorithm for MAT}
\label{alg:Framwork}
\begin{algorithmic}[1]
\For{number of training iterations}
\State Sample a batch of labeled data $(x_l, y_l)$ with size $N_l$ and a batch of unlabeled data $(x_{ul})$ with size $N_{ul}$. $Z_{l_i}$ denotes the matrix of latent features of data with label $i$ in a batch. $\Sigma_{i}$ and $\mu_{i}$ denote the covariance matrix and mean for $i^{th}$ component of Gaussian Mixtures.
    \For{j in 1...n}
    \State $d\gets\overline{\nabla_{\epsilon}M_{KL}(\epsilon, x, \theta)|_{\epsilon=\xi d}}$
    \EndFor
\State $\epsilon_{adv} = \xi d$
\State Update the parameters of neural network with stochastic gradient: \\
\State $-\nabla_{\theta}\{\frac{1}{N}\sum_{i=1}^{N} log P(y_i|x_i, \theta) - \lambda_1\frac{1}{N}\sum_{i=1}^{N} log S_{kl}(x_i, \theta) + \lambda_2I_{\theta}\}$
\State Update the covariance matrix and mean of Gaussian Mixtures using the latent features:
    \For{i in 1...m}
    \State $\Sigma_{i}^{t+1}=\alpha\Sigma_{i}^{t} + (1-\alpha)\frac{1}{N_{l_i}}X_{l_i}^{T}X_{l_i}$
    \State $\mu_{i}=\beta\mu_{i} + (1-\beta)mean(X_{l_i})$
    \EndFor
\EndFor
\end{algorithmic}
\end{algorithm}

\subsection{Computational Analysis}
  We briefly touch the computational analysis here. Compared with previous methods like VAT~\cite{miyato2017virtual}, our proposed method needs to compute the mean and covariance matrix for Gaussian mixtures additionally. In this work, parameters of Gaussian mixtures are evaluated by the latent features of labeled data instead of stochastic gradient. It may be inefficient to compute the inverse of covariance matrix. However, in the low-dimensional latent space, the inverse of covariance matrix can be easily and directly obtained; even in the very high-dimensional space,  the covariance matrix can be assumed as a diagonal matrix which is also easy to get its inverse. Same as VAT, we just need one iteration of the power method to estimate the worst perturbation $\epsilon_{adv}$. The whole procedure for updating the parameters of deep neural networks consists of two forward  and two back propagations. The first forward and backward propagations are used to evaluate the worst perturbation. After calculating the final loss, propagate forward and backward again to update the parameters of neural network and Gaussian mixtures.

\section{Experiment}
For assessing the efficacy of our proposed MAT, we implemented it on several benchmark datasets including MNIST, CIFAR-10, and SVHN.  In principle, adversarial methods can be regarded as a robust method which typically have better generalization abilities than the traditional methods. To check if the proposed MAT can indeed improve the classification performance, we first applied MAT on the on benign data  of MNIST and CIFAR-10 in both the supervised and semi-supervised task. Additionally, to check further the robustness of our proposed adversarial training framework, we perform various experiments to examine how the proposed MAT could defend the attack from various adversarial methods. To illustrate our proposed method, a series of visualizations were made, offering some interesting results which might be used to explain the adversarial examples.

\subsection{Experiments on Benign Data}
We first evaluated the performance of our proposed MAT methods against many other comparison methods on the benign data. We first report the performance of various methods in the setting of supervised classification and then perform the  comparison in semi-supervised learning.
\subsubsection{Supervised Learning}
We implement the same framework Lenet++ with~\cite{wen2016discriminative} on MNIST dataset. For Lenet++, there are only two dimensions in the last hidden layer which is convenient for visualization. The based framework for experiment on CIFAR-10 is the same as~\cite{miyato2017virtual} called Conv-Large. For MNIST dataset, we train the deep model with $60,000$ labeled training samples, and we evaluate it with $10,000$ test samples. For CIFAR-10, we use $50,000$ training samples and $10,000$ test samples. To search good hyper parameters, the training set is divided into $50,000$ training set and $10,000$ validation set.  We choose a set of hyper parameters with the best performance on the validation set. For MNIST dataset, the best hyper parameters are obtained as: $\lambda_1=0.1$, $\lambda_2=0.1$ and $\sigma=2$. For CIFAR-10, the parameters are obtained as: $\lambda_1=0.01$, $\lambda_2=0.1$ and $\sigma=20$.

\begin{table}[h!]
  \caption{Test performance on MNIST in supervised learning}
  \label{sup-table1}
  \centering
  \begin{tabular}{ll}
    \hline
    \multirow{2}{2cm}{Method}      & MNIST  \\
                                    & Test error rate($\%$) \\
    \hline
    SVM  & $1.40$     \\
    Dropout~\cite{srivastava2014dropout}      & $1.05$       \\
    Ladder networks~\cite{rasmus2015semi}    & $0.57\pm 0.02$  \\
    Adversarial, $L_{\infty}$ norm constraint~\cite{goodfellow2014explaining}   & $0.78$     \\
    Adversarial, $L_2$ norm constraint~\cite{miyato2017virtual}  & $0.71$      \\
    RPT~\cite{miyato2017virtual}  & $0.82$     \\
    \hline
    Baseline    &  $1.10\pm 0.03$   \\
    Center loss &  $0.86\pm 0.018$   \\
    VAT  & $0.72\pm 0.016$      \\
    MAT  & $\textbf{0.42}\pm 0.021$     \\
    \hline
  \end{tabular}
\end{table}

\begin{table}[h!]

  \caption{Test performance on CIFAR-10 in supervised learning}
  \label{sup-table2}
  \centering
  \begin{tabular}{ll}
    \hline
    \multirow{2}{2cm}{Method}      & CIFAR-10  \\
                                    & Test error rate($\%$) \\
    \hline
    Network in Network~\cite
    {lin2013network}  & $8.81$     \\
    All-CNN~\cite{springenberg2014striving}      & $7.25$       \\
    Deeply Supervised Net~\cite{lee2015deeply}    & $7.97$  \\
    Highway Network~\cite{srivastava2015highway}   & $7.72$     \\
    RPT~\cite{miyato2017virtual}  & $6.25\pm 0.04$     \\
    Baseline    &  $6.76\pm 0.07$   \\
    VAT  & $5.81\pm 0.02$      \\
    \hline
    MAT  & $\textbf{4.40}\pm 0.03$     \\
    \hline
  \end{tabular}
\end{table}

Table~\ref{sup-table1} and Table~\ref{sup-table2} list the performance of our proposed method and other competitive methods on MNIST and CIFAR-10. Generally, adversarial methods can be regarded as a robust method which typically have better generalization abilities than the traditional methods. As observed from both the tables,  MAT clearly achieves the best performance among all the comparison methods in both the datasets. It is probably unfair to compare with some methods, e.g., ladder network, simply because the different based model were used there. However, it is sufficient to show the  superiority of our proposed MAT due to its significant lower error rate than most of the other competitive adversarial methods.


\subsubsection{Semi-supervised Learning}
In the methodology part, we have introduced the MAT's final objective function Eq.~(\ref{eq:Final}) where the first term is the ordinary soft-max loss function and the last term is the information regularization term. These two terms are exploited on the training data with labels. And the second term is used to smooth the manifold and output distribution which does not need the label information. Therefore, our method can readily be extended to semi-supervised learning. Following the same setting as~\cite{miyato2017virtual}, we implement our MAT both on CIFAR-10 and SVHN with $4,000$ labeled data and $1,000$ labeled data respectively. We use the same base model with~\cite{miyato2017virtual} called Cov-Large with batch normalization and dropout. We exploit the mini-batch of size $32$ for both labeled data and unlabeled data on CIFAR-10. For SVHN, we use the labeled batch with size $32$ and unlabeled batch with size $128$.

\begin{table}[h!]

  \caption{Test performance on SVHN (1,000 labeled)}
  \label{semi-table1}
  \centering
  \begin{tabular}{ll}
    \hline
    \multirow{2}{2cm}{Method}      & SVHN  \\
                                    & Test error rate($\%$) \\
    \hline
    SWWAE~\cite{zhao1506stacked}  & $23.56$      \\
    Skip Generative Model~\cite{maaloe2016auxiliary}      & $16.30\pm 0.24$        \\

    GAN with feature matching~\cite{salimans2016improved}  & $8.11\pm 1.3$      \\
    $\Pi$ model~\cite{laine2016temporal}  & $5.43$      \\
    RPT~\cite{miyato2017virtual}  & $8.41\pm 0.24$  \\
    \hline
    VAT  & $5.77\pm 0.32$     \\
    MAT  & $\textbf{4.90}\pm 0.17$  \\
    \hline
  \end{tabular}
\end{table}

\begin{table}[h!]
  \caption{Test performance on CIFAR-10 (4,000 labeled)}
  \label{semi-table2}
  \centering
  \begin{tabular}{ll}
    \hline
    \multirow{2}{2cm}{Method}      & CIFAR-10  \\
                                    & Test error rate($\%$) \\
    \hline

    Ladder networks, $\Gamma$ model~\cite{rasmus2015semi}   & $20.40\pm 0.47$  \\
    CatGAN~\cite{springenberg2015unsupervised}   & $19.58\pm 0.58$     \\
    GAN with feature matching~\cite{salimans2016improved}   & $18.63\pm 2.32$     \\
    $\Pi$ model~\cite{laine2016temporal}  & $16.55\pm 0.29$     \\
    RPT~\cite{miyato2017virtual}
    & $18.56\pm 0.29$     \\
    \hline
    VAT   & $14.82\pm 0.13$     \\
    MAT   & $\textbf{12.85}\pm 0.21$     \\
    \hline
  \end{tabular}
\end{table}

Table~\ref{semi-table1} and Table~\ref{semi-table2} show the performance of our proposed model as well as recent state-of-the-art semi-supervised learning methods on CIFAR-10 and SVHN. Once again, our proposed method MAT achieves the highest performance, which further shows its advantages.

\subsubsection{Visualization}
In order to illustrate why the proposed MAT could perform excellent, we take the MNIST as one example to visualize various methods. In particular, Figure~\ref{tsnetext} shows the illustration of the last embedding space of Lenet++ with center loss, softmax, VAT, and MAT on the MNIST test data.\footnote{Note that the graph has been smoothed with fitting a Gaussian over each point in order to generate clear visualizations.} It is obvious that the latent features of our method are represented more compactly and discriminatively with respect to class centers. VAT learns the similar latent space with softmax, but apparently both of their features are not as discriminative as our proposed MAT. This may illustrate why our proposed MAT method could usually generate better classification performance than VAT and the traditional CNN.

\begin{figure}[h!]
\label{fig3}
\centering
\subfigure[Learning curves for CIFAR-10.] {
\psfig{file=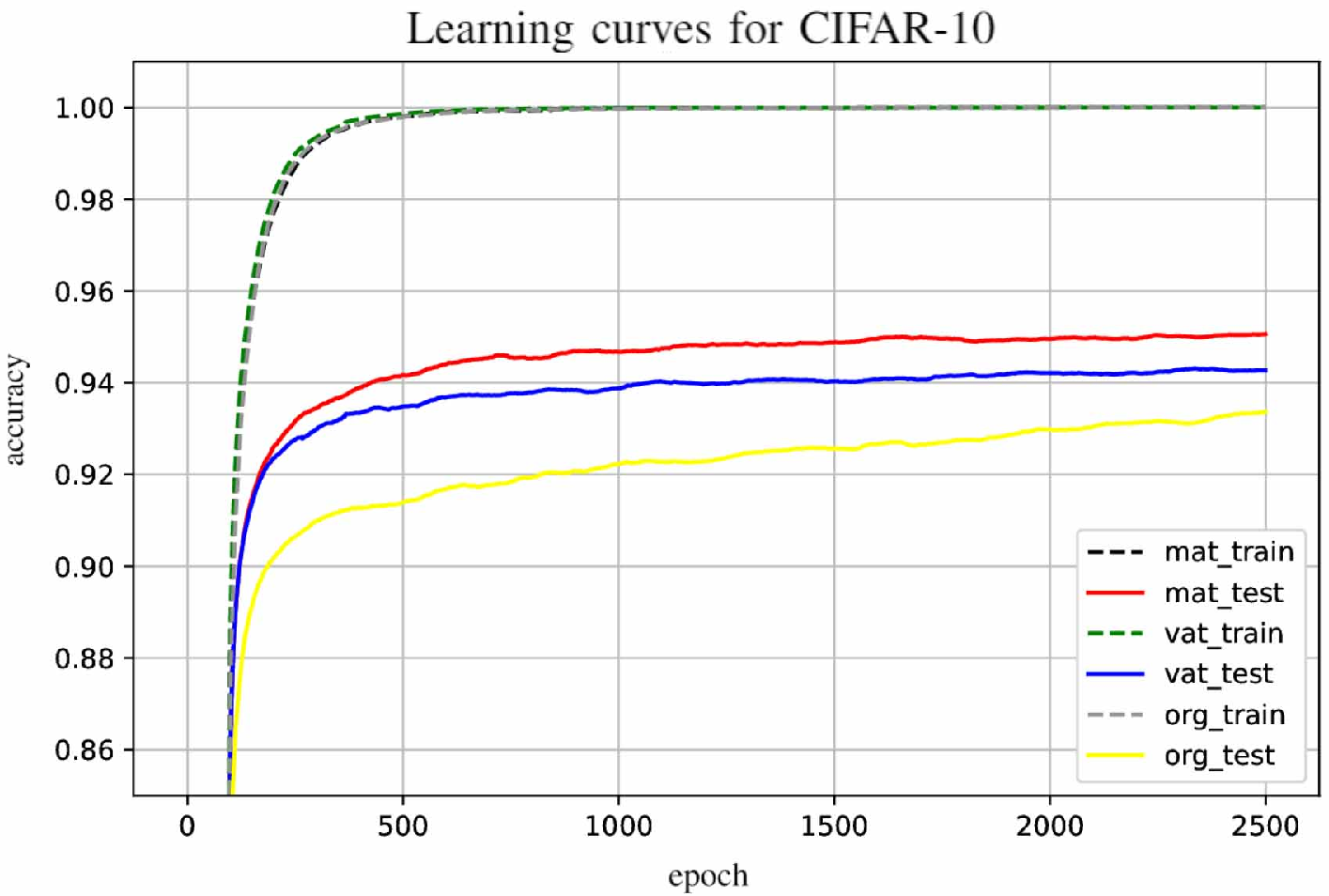,angle=0,width=0.47\textwidth}
\label{fig:tsne5} }
\subfigure[Smoothness on CIFAR-10] {
\psfig{file=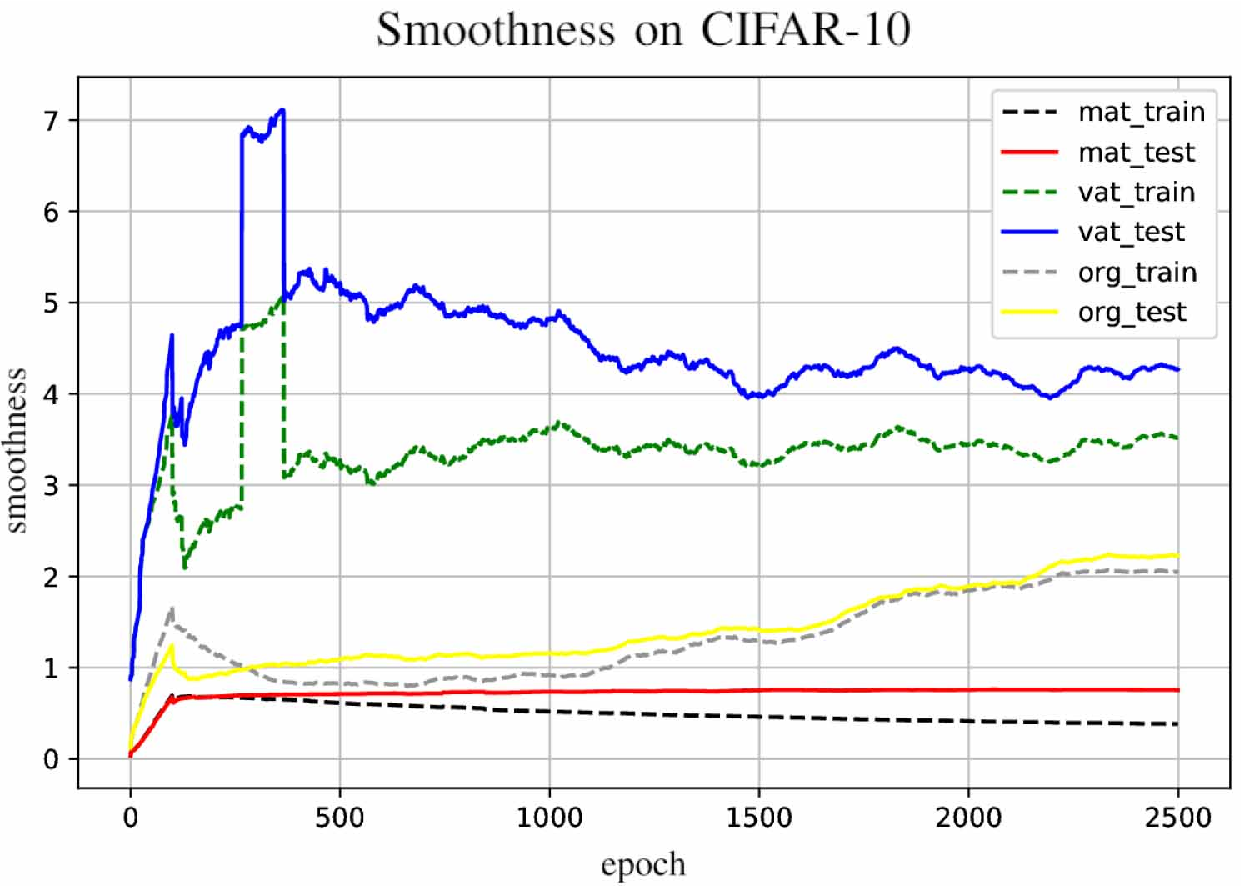,angle=0,width=0.45\textwidth}
\label{fig:tsne6} }
\caption{(a) The learning curve of both the training and test on CIFAR-10 (recognition rate). (b) The smoothness for our proposed MAT and two comparatives (VAT and traditional CNN) The graphs are better viewed in color.}\label{curve}
\end{figure}

Since our proposed MAT seeks a smooth latent space  which we believe would benefit the classification, we plot in Figure~\ref{curve} the learning curve (accuracy rate) and the smoothness for the three different methods: MAT (our proposed method), VAT, and the baseline (traditional CNN) on CIFAR-10.\footnote{Our proposed MAT and VAT are both extended with the same baseline model.} For MAT,  the best set of hyperparameters, i.e., $\lambda_1=0.01$  $\lambda_2=0.1$ and $\sigma=20$ was used, while for VAT, the best setting reported in~\cite{miyato2017virtual} was directly applied. Compared with the other two methods, our proposed MAT clearly increases the smoothness $R_{smooth}$, which is defined by the average of the smoothness of output space and latent space:
\begin{eqnarray}
\begin{aligned}
R_{smooth} =& \frac{1}{2N}\sum_{i=1}^{N} KL[P(y|x,\theta)||P(y|x+\epsilon,\theta)]
\\ +&KL[P_{GM}(f(x,\theta))||P_{GM}(f(x+\epsilon,\theta))]
\label{eq:10}
\end{aligned}
\end{eqnarray}
When we calculate the smoothness, the perturbation needs to be normalized to unit vector. Figure~\ref{curve}(b) shows that our proposed method indeed learns the smoother latent and output space.

\subsection{Performance on Defending Adversarial Examples}
We now turn to examining how the proposed MAT method could defend the attacks from various adversarial examples generation approaches in comparison with  other competitive methods. Visualization is also presented so as to obtain further understandings on adversarial examples.

\subsubsection{Robustness to Adversarial Attacks}
\begin{figure*}[h!]
\label{fig45}
\centering
\subfigure [FGSM attack on MNIST. ]{
\psfig{file=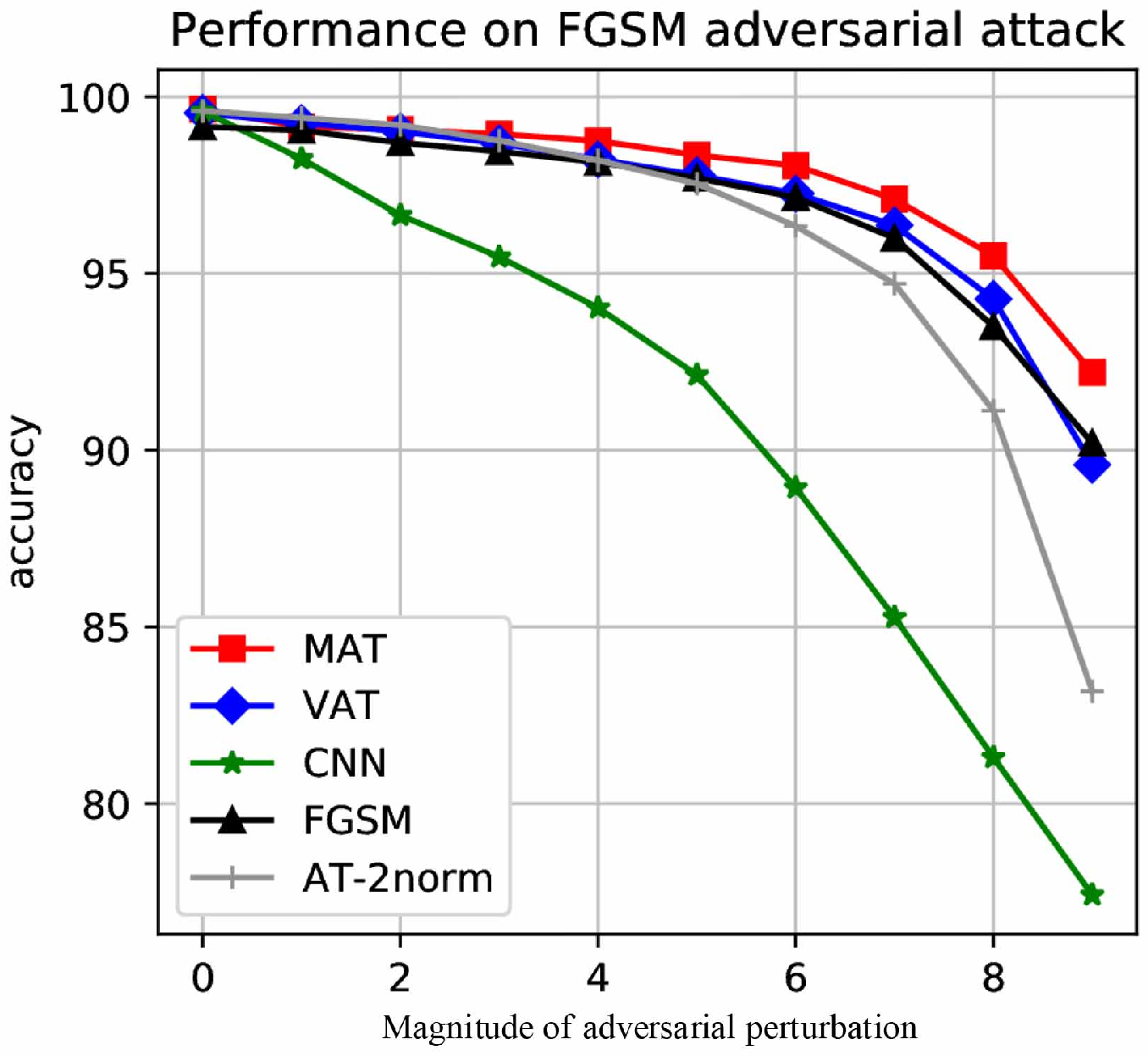,angle=0,width=0.3\textwidth}
\label{fig:ad1} }
\subfigure [FGSM attack on CIFAR-10. ]{
\psfig{file=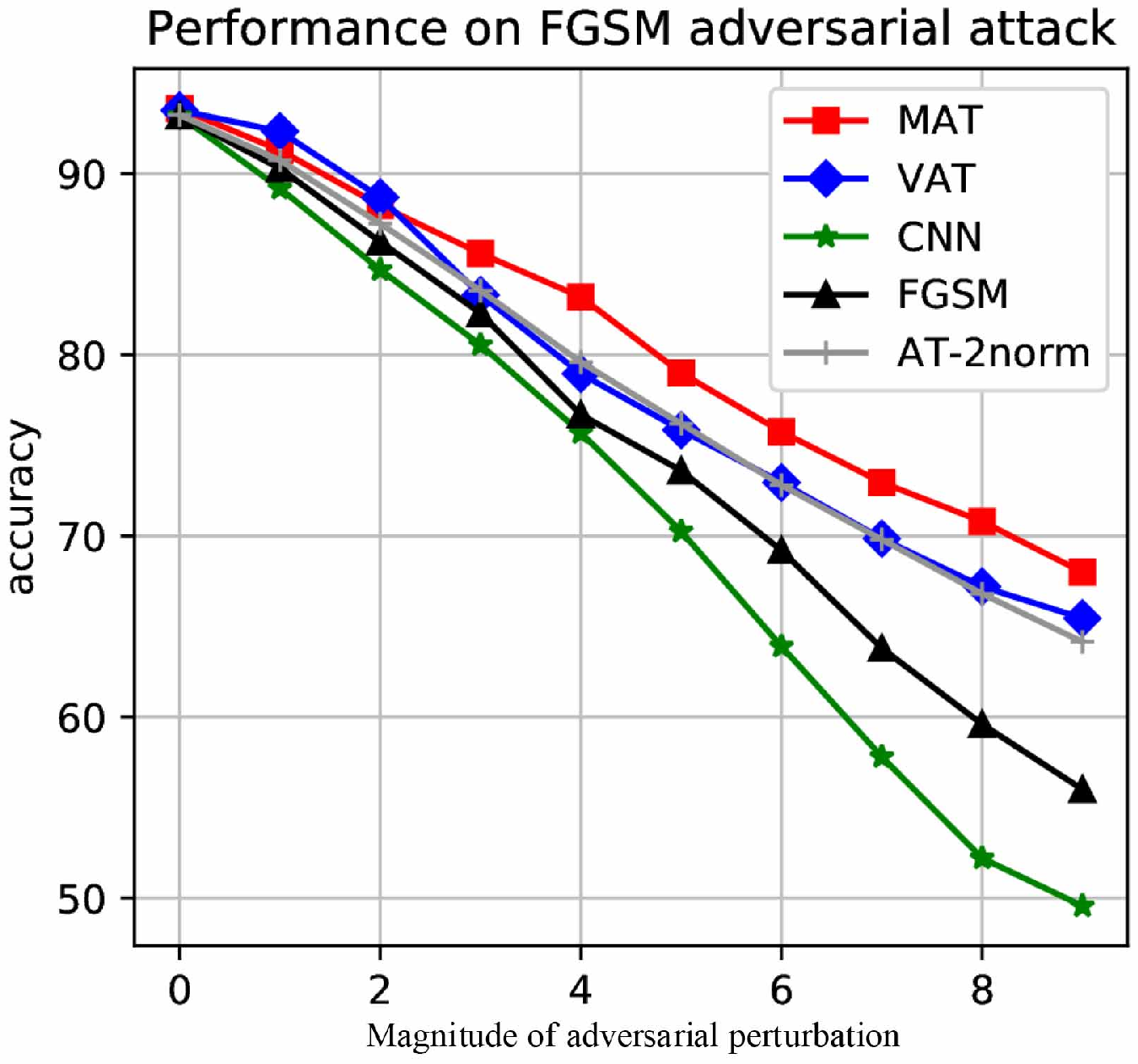,angle=0,width=0.3\textwidth}
\label{fig:ad2} }
\subfigure [FGSM attack on SVHN ]{
\psfig{file=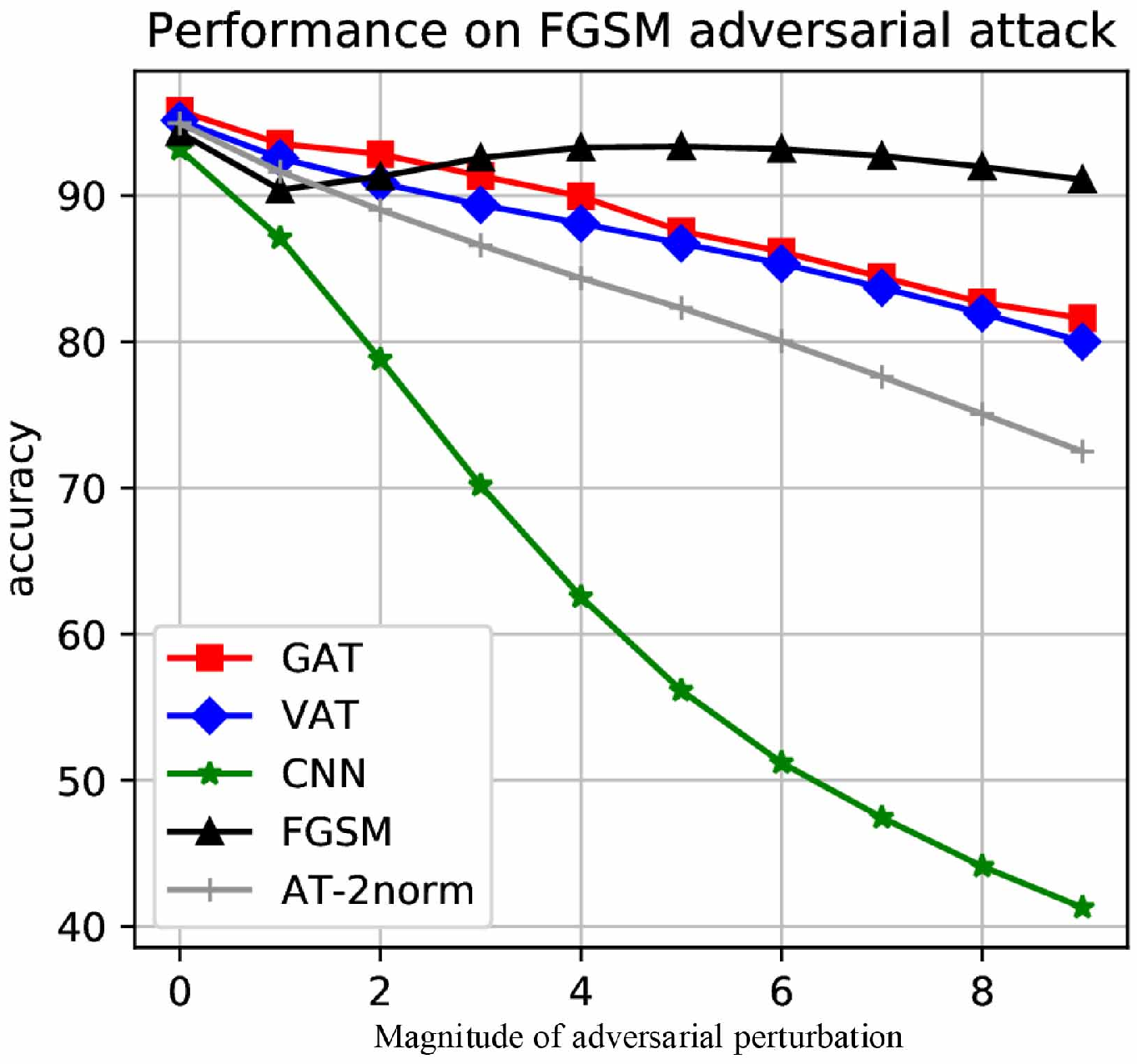,angle=0,width=0.3\textwidth}
\label{fig:ad3} }
\subfigure [2-norm attack on MNIST. ]{
\psfig{file=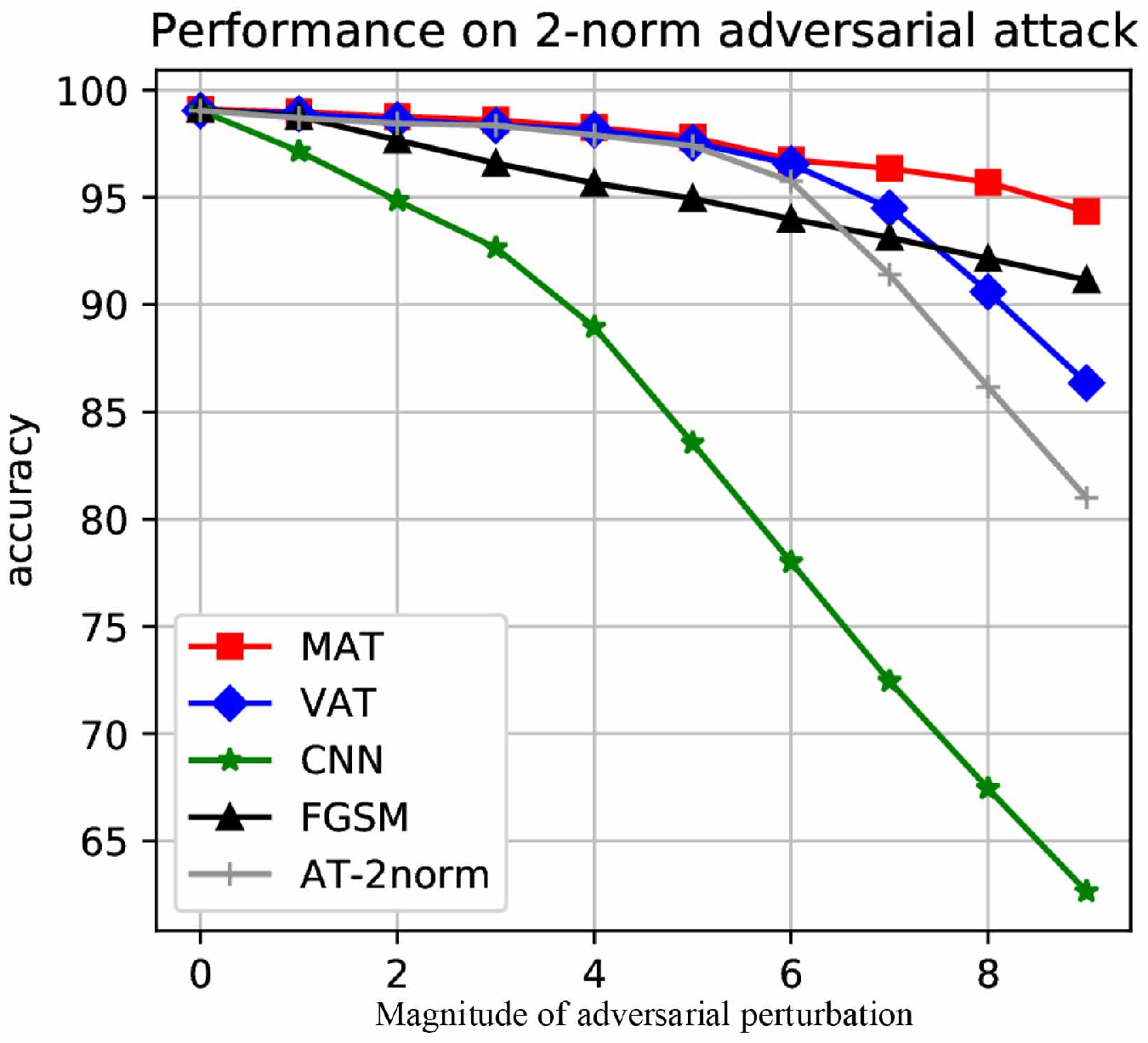,angle=0,width=0.3\textwidth}
\label{fig:ad4} }
\subfigure [2-norm attack on CIFAR-10. ]{
\psfig{file=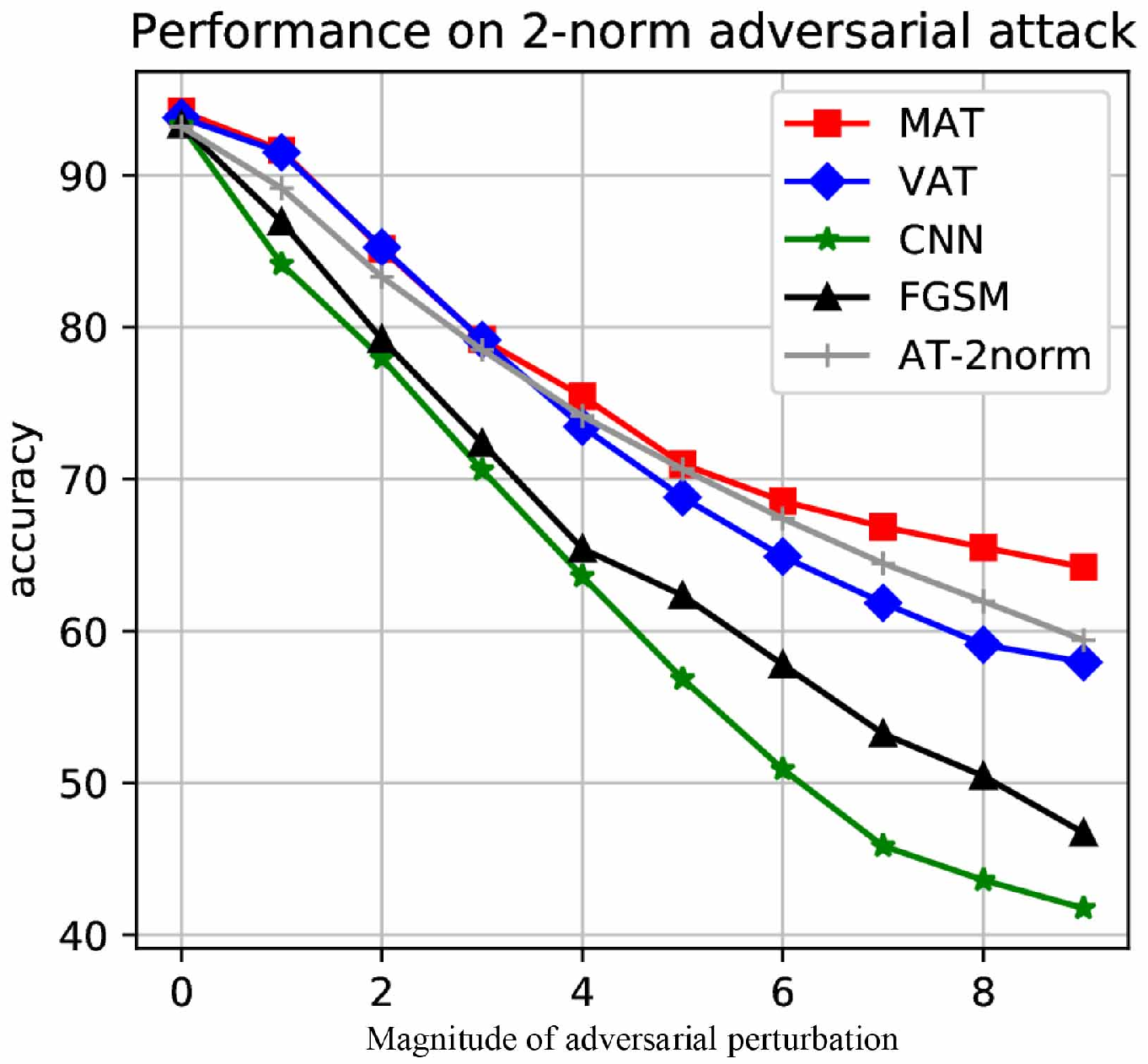,angle=0,width=0.3\textwidth}
\label{fig:ad5} }
\subfigure [2-norm attack on SVHN. ]{
\psfig{file=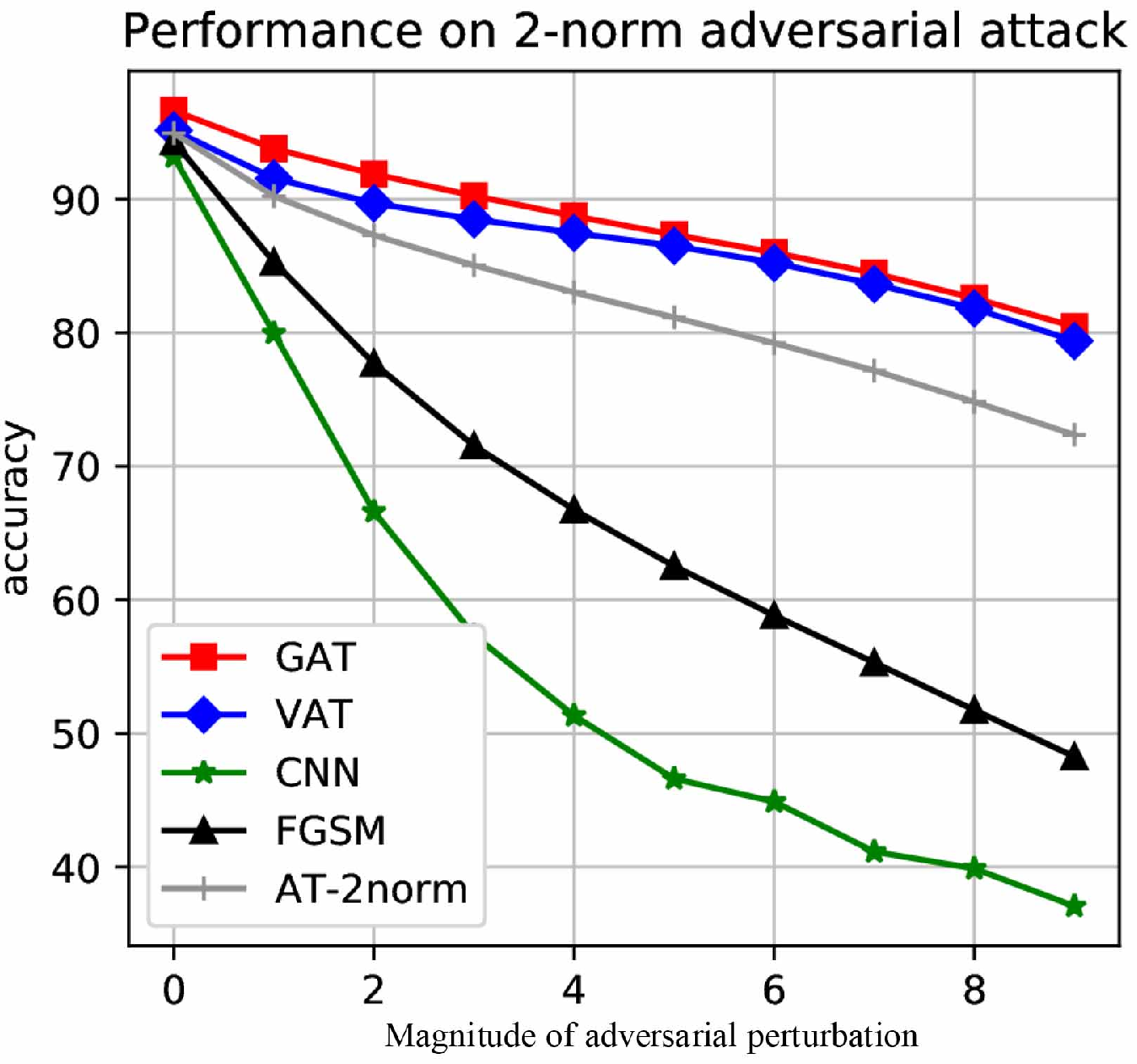,angle=0,width=0.3\textwidth}
\label{fig:ad6} }
\caption{(a)-(c) Performance of different methods on FGSM attack. (d)-(f) Performance of different methods on 2-norm adversarial attack.}\label{visual_per}
\end{figure*}

\begin{figure*}[h!]
\label{fig45}
\centering
\subfigure [MAT ($\sigma=0$). ]{
\psfig{file=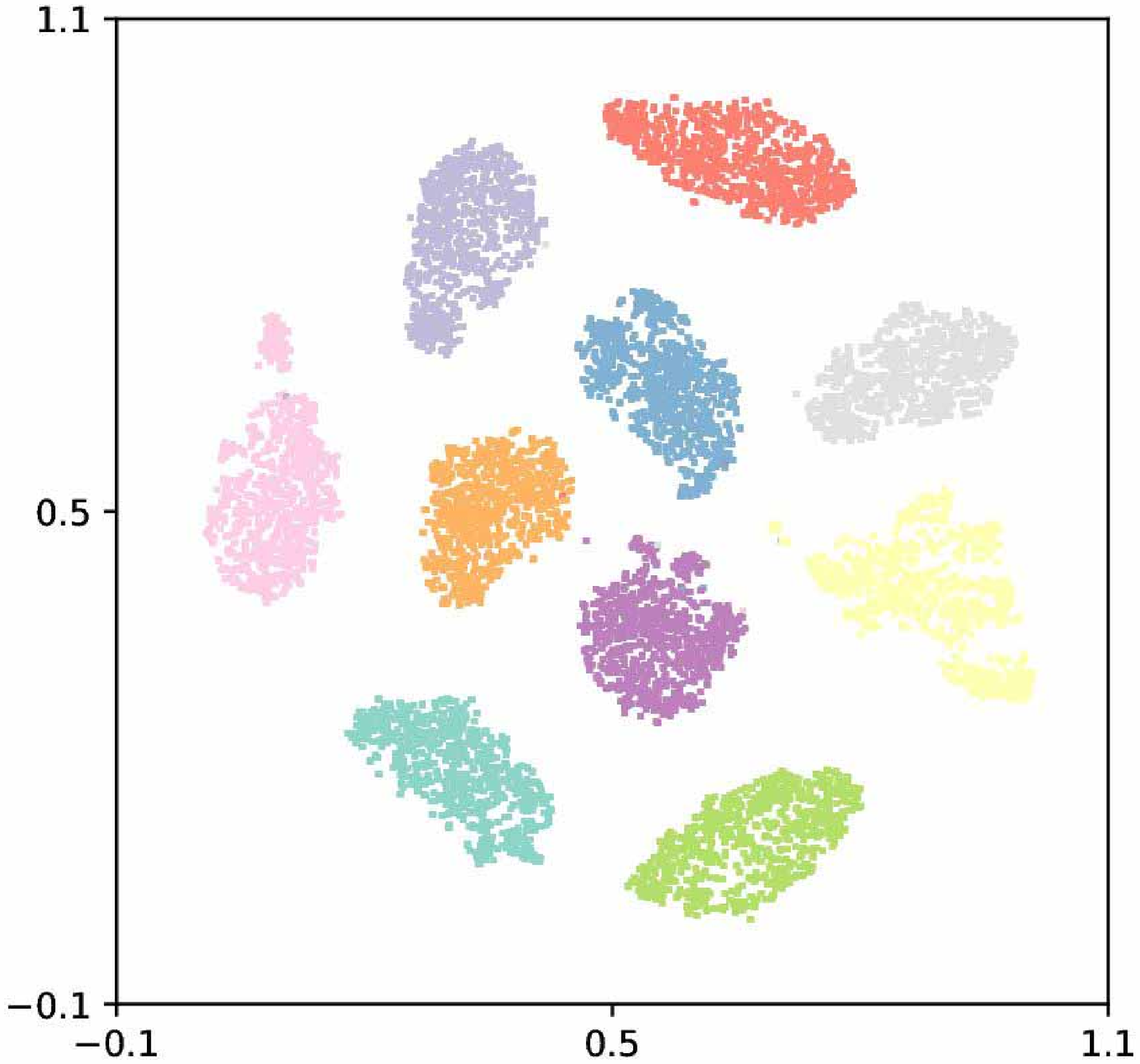,angle=0,width=0.23\textwidth}
\label{fig:lad1} }
\subfigure [MAT ($\sigma=2$). ]{
\psfig{file=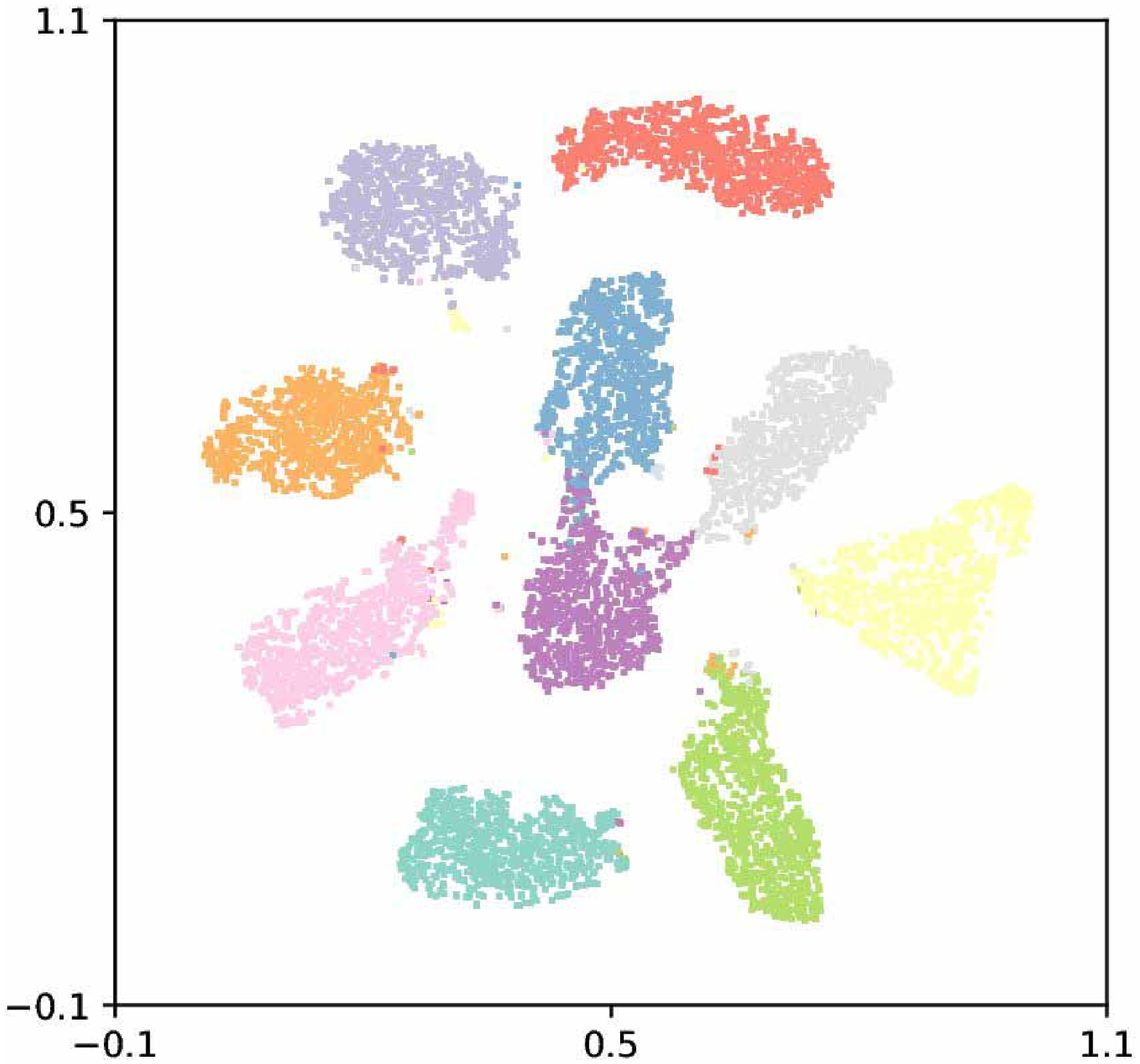,angle=0,width=0.23\textwidth}
\label{fig:lad2} }
\subfigure [MAT ($\sigma=4$). ]{
\psfig{file=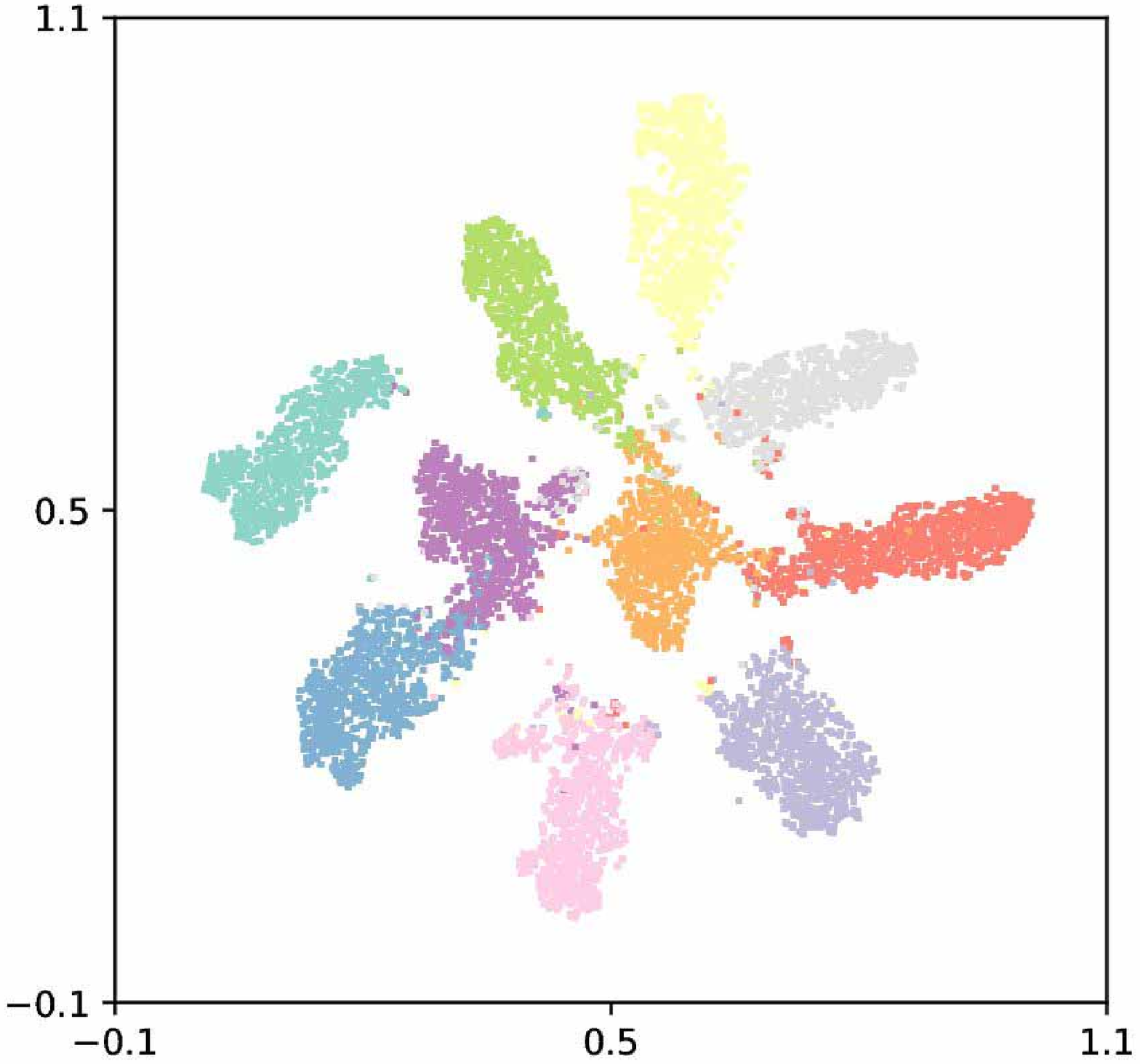,angle=0,width=0.23\textwidth}
\label{fig:lad3} }
\subfigure [MAT ($\sigma=6$). ]{
\psfig{file=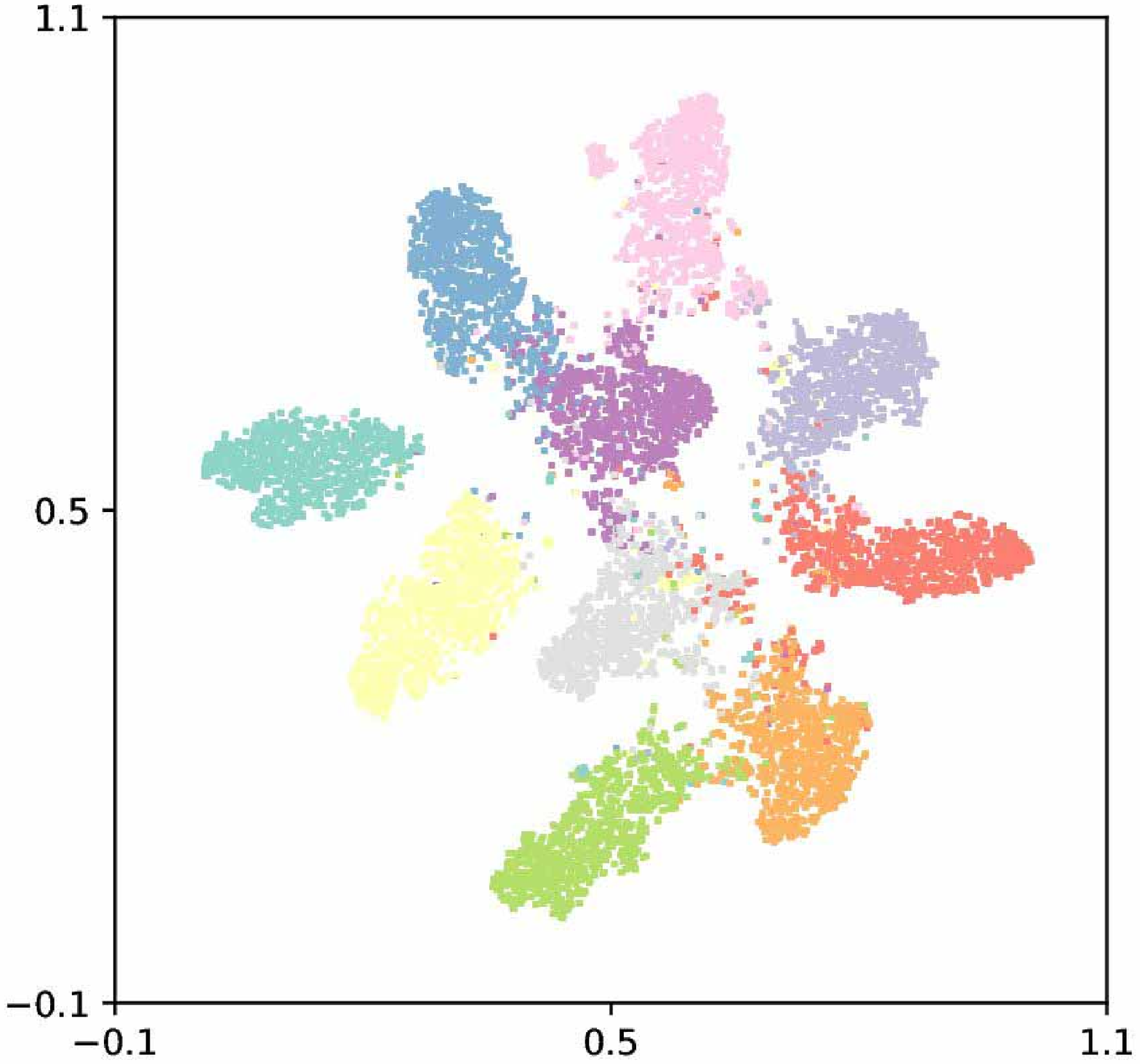,angle=0,width=0.23\textwidth}
\label{fig:lad4} }
\subfigure [VAT ($\sigma=0$). ]{
\psfig{file=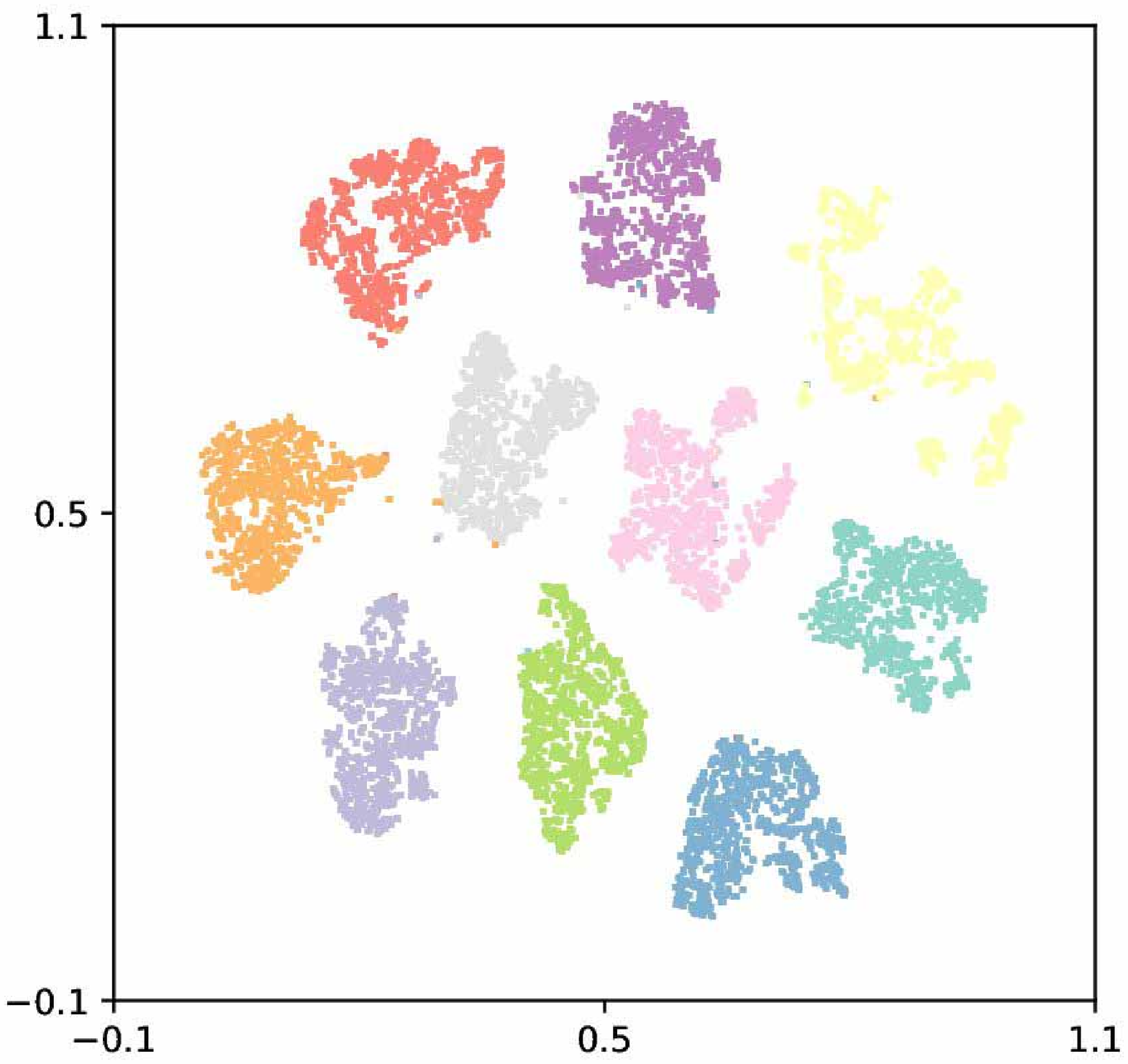,angle=0,width=0.23\textwidth}
\label{fig:lad5} }
\subfigure [VAT ($\sigma=2$). ]{
\psfig{file=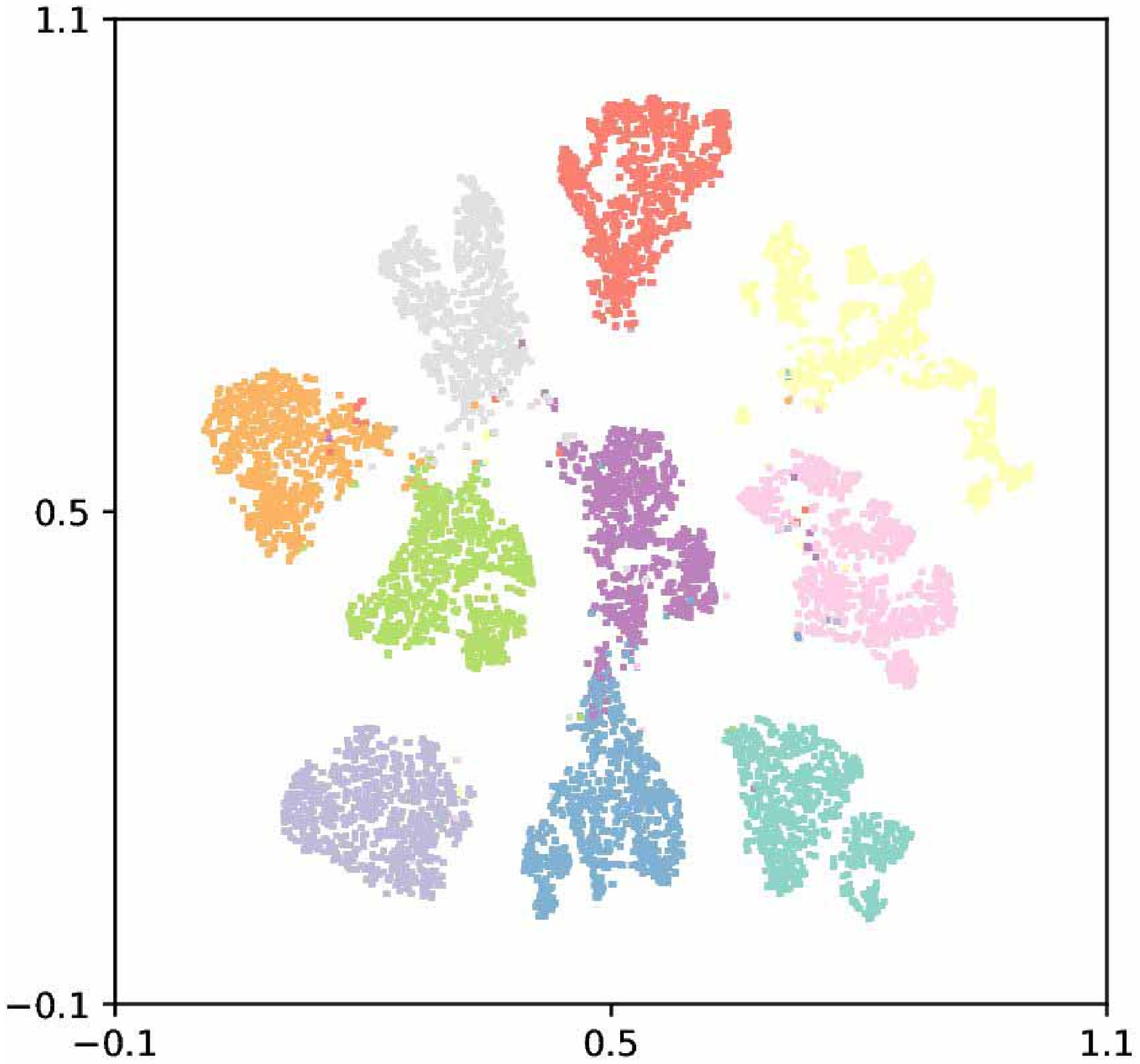,angle=0,width=0.23\textwidth}
\label{fig:1ad6} }
\subfigure [VAT ($\sigma=4$). ]{
\psfig{file=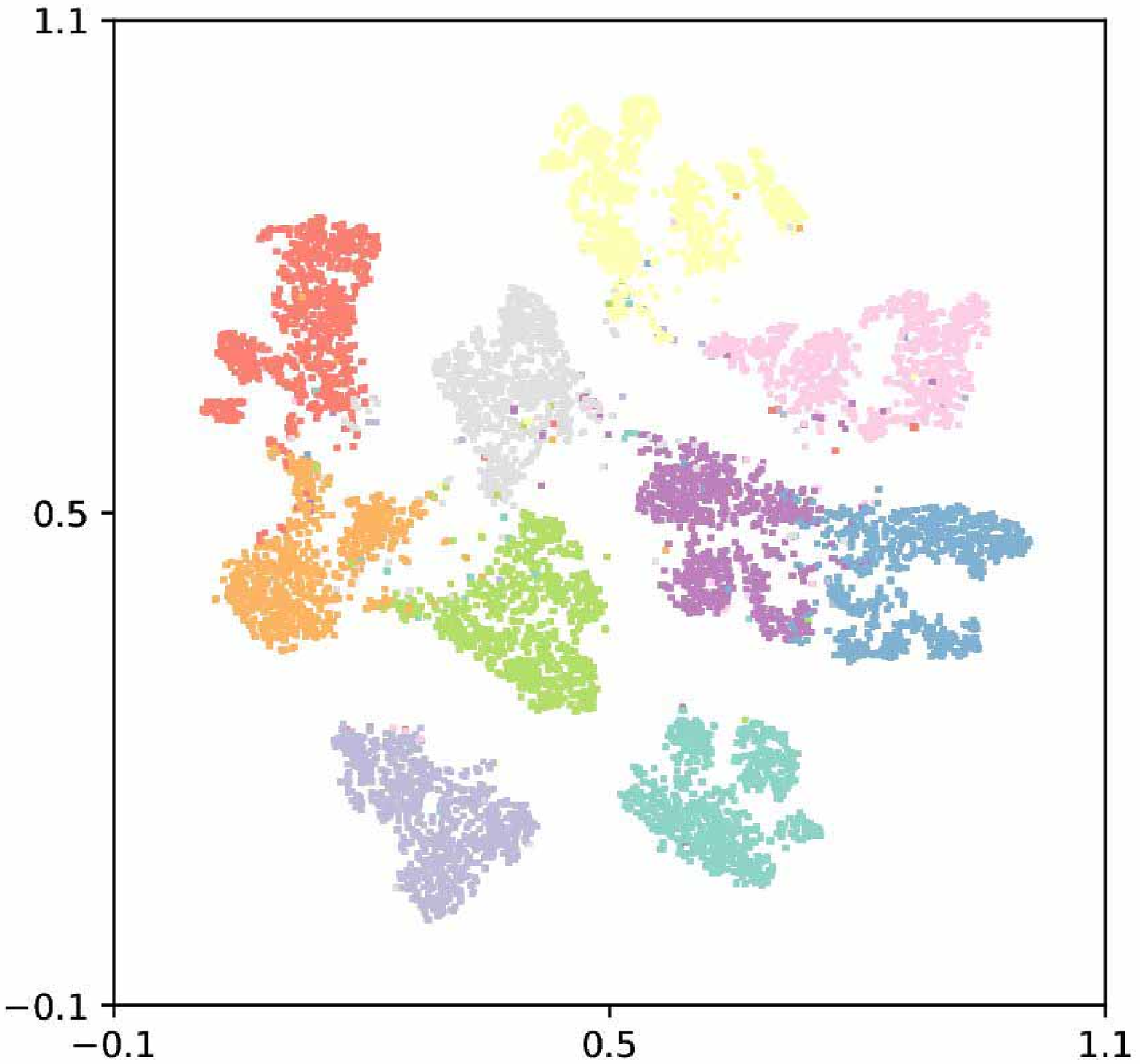,angle=0,width=0.23\textwidth}
\label{fig:1ad7} }
\subfigure [VAT ($\sigma=6$). ]{
\psfig{file=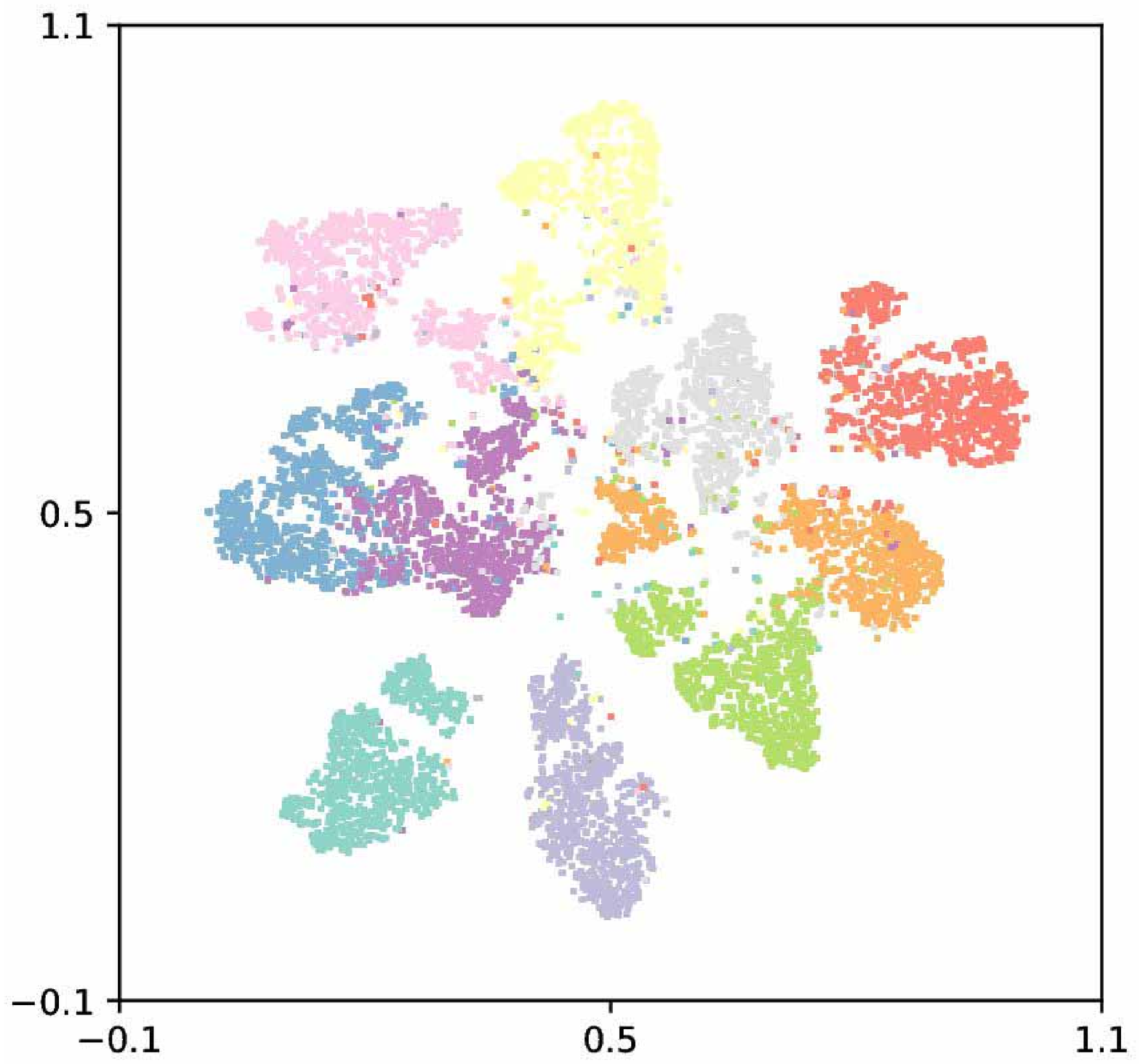,angle=0,width=0.23\textwidth}
\label{fig:lad8} }
\caption{(a)-(d) The TSNE embedding for the latent space of MAT on 2-norm adversarial attack with different $\sigma$  (a)-(d) The TSNE embedding for the latent space of VAT on 2-norm adversarial attack with different $\sigma$}\label{visual_TSNE}
\end{figure*}

We implement different methods on FGSM and 2-norm adversarial attack for MNIST, CIFAR-10, and SVHN datasets. Particularly, we generate in the test sets of MNIST and CIFAR-10 10,000 adversarial examples according to FSGM and 2-norm attacks~\cite{lyu2015unified} respectively. For SVHN, we generate 26,032 adversarial examples. We increase the level of adversarial noise gradually from 0 to 8 in MNIST and SVHN with the step size as 1 and from 0 to 13 in CIFAR-10 with the step size as 1.6. We then test the performance of various training methods on these adversarial examples. The performance is plotted in Fig.~\ref{visual_per}. As clearly observed, the proposed MAT shows better robustness against the two types of adversarial examples. Particularly, when the adversarial noises are small, all the adversarial training methods show similar results but perform much better than the CNN (exploiting no adversarial training); when the adversarial attacks are stronger,  the proposed method overall demonstrates clearly better performance in almost all the cases, verifying its significant robustness. One exception can be identified in Fig.~\ref{visual_per}(c) where the FGSM training method performs the best while our proposed MAT performs the second. However, since the FGSM training was specifically designed to defend the FGSM adversarial examples, it will not be very surprising that FGSM can do better against the FGSM adversarial examples. Nonetheless, except Fig.~\ref{visual_per}(c), our MAT still demonstrates the best robustness in all the other cases.

Again, taking MNIST as one example, we show the TSNE embedding for the second last layer given by VAT and MAT under various levels of 2-norm adversarial attack. As Figure~\ref{visual_TSNE} shows, when the degree of adversarial noise increases (from 0 to 6 with step size 2), the clusters of different classes are less discriminative. In contrast to VAT, our proposed MAT obtains much more discriminative features on different levels of adversarial perturbation.
\begin{figure*}[h!]
\label{fig45}
\centering
\subfigure {
\psfig{file=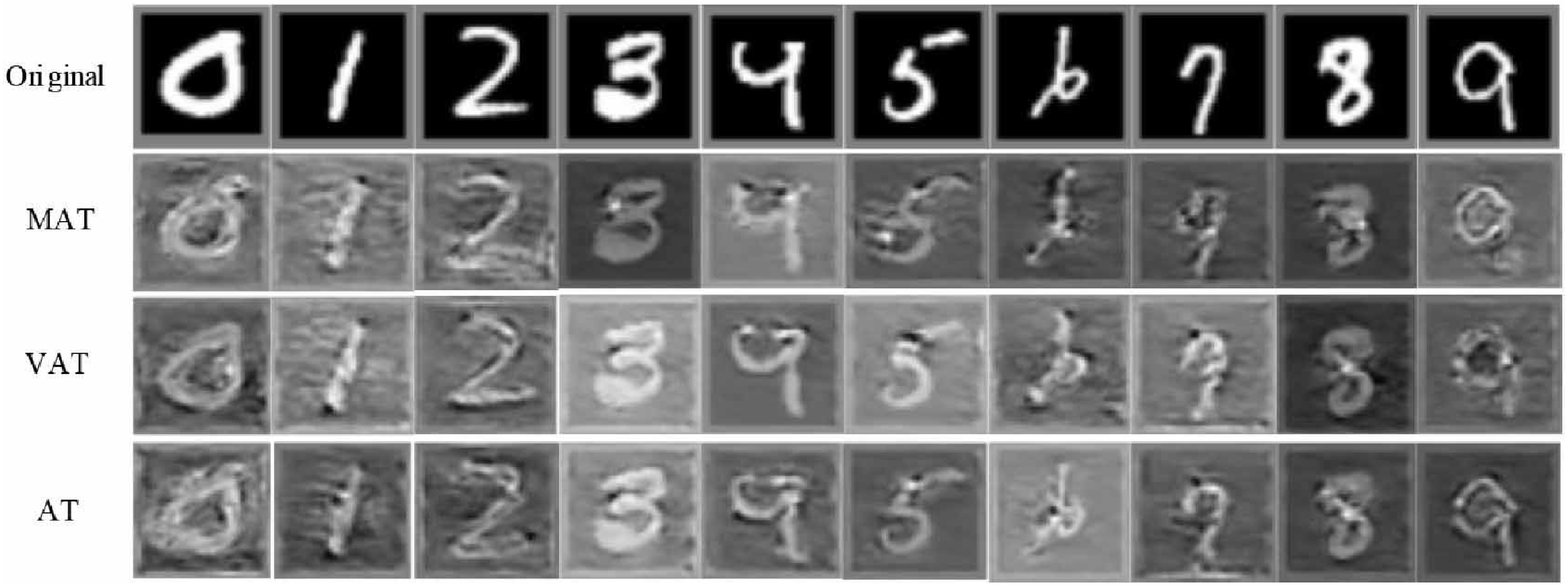,angle=0,width=0.8\textwidth}
\label{fig:tsne5} }
\caption{Adversarial examples generated on MNIST by different methods}\label{visual_m}
\end{figure*}

\begin{figure*}[h!]
\label{fig45}
\centering
\subfigure {
\psfig{file=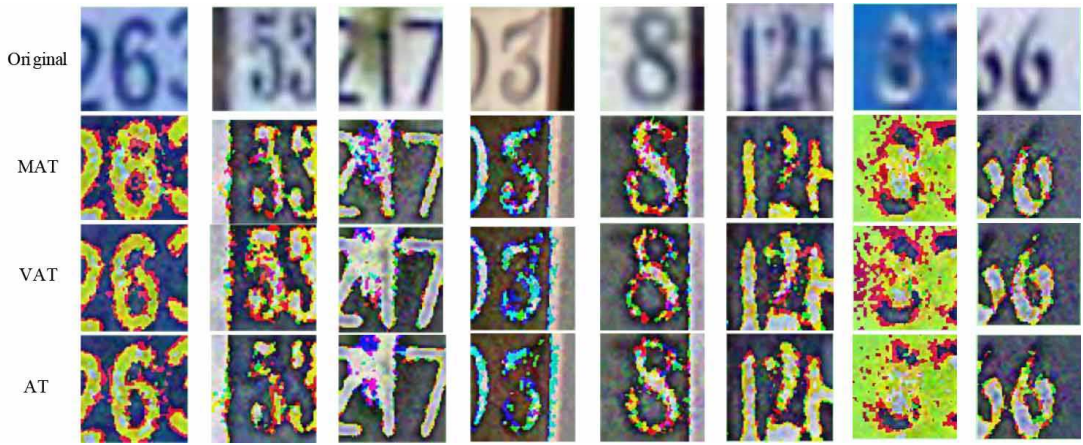,angle=0,width=0.8\textwidth}
\label{fig:tsne5} }
\caption{Adversarial examples  generated on SVHN by different methods}\label{visual_s}
\end{figure*}

\subsubsection{Visualization of Adversarial Examples}
\begin{figure*}[h!]
\label{fig45}
\centering
\subfigure {
\psfig{file=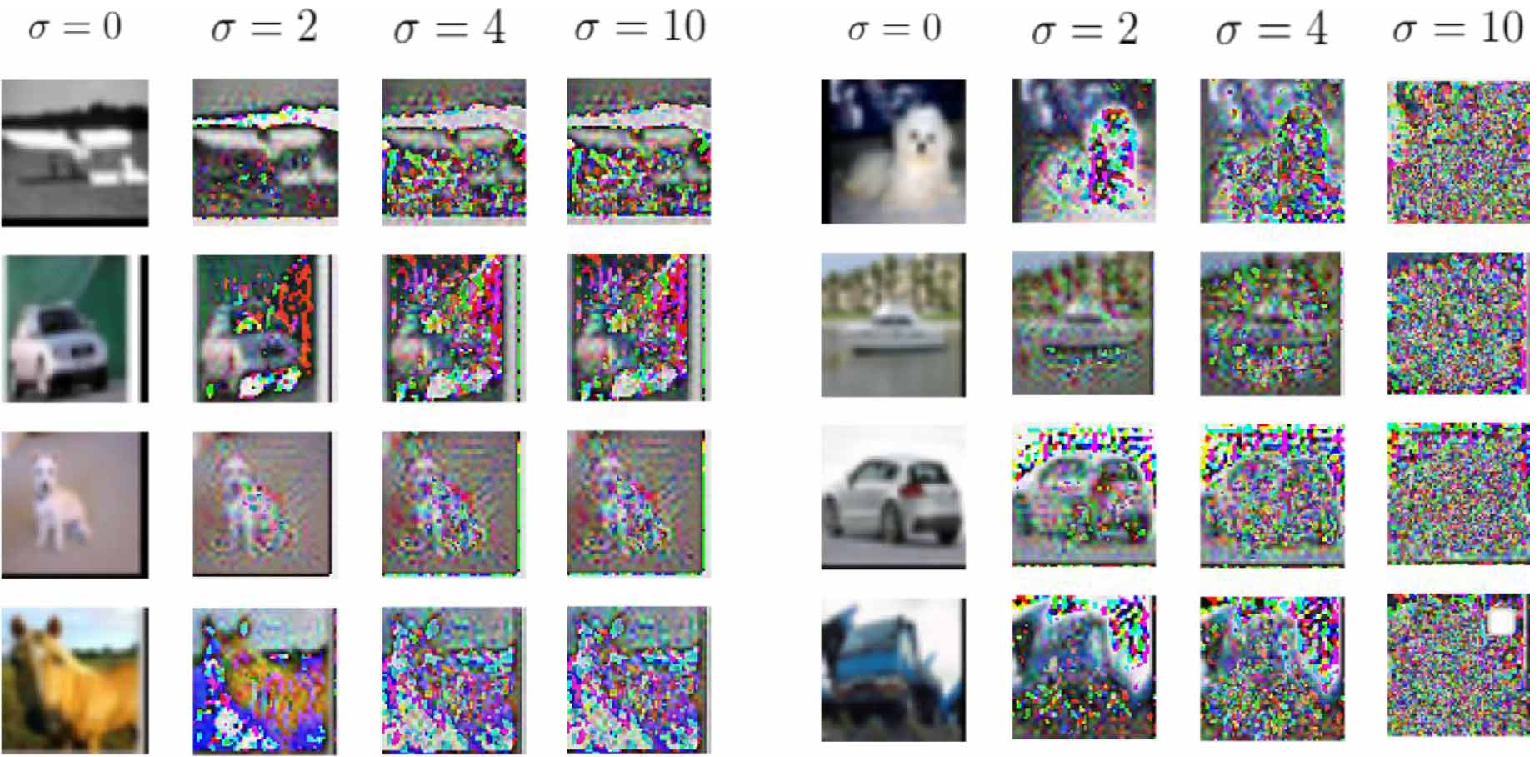,angle=0,width=0.8\textwidth}
\label{fig:tsne5} }
\caption{Adversarial examples  generated on  CIFAR-10 by different methods}\label{visual_c}
\end{figure*}

In this subsection, we visualize some adversarial examples generated by both our proposed method and the other comparison adversarial methods. To visualize the adversarial examples, we directly train the deep neural network with original images and then calculate the worst perturbation through back propagation. For better visualization, the image up-sampling was applied.

Specifically, we  compare the adversarial examples generated by our method (MAT), VAT, and traditional adversarial learning on MNIST, SVHN, and CIFAR10. Figure~\ref{visual_m}-\ref{visual_c} show several adversarial examples (generated by different methods) from the three datasets. It is interesting to note that after the adversarial training given by MAT,  the original images tend to be morphed into some other categories. Particularly, digit "$3$" was changed into "$8$", "$7$" was changed to "$9$", and "$9$" was changed to "$0$" on MNIST (see Figure~\ref{visual_m}). Similar cases can also be observed in Figure~\ref{visual_s} where the worst perturbation generated by MAT tends to change the number "$6$" to number "$8$" (the 1st image in the 2nd row of Figure~\ref{visual1}), the number "$3$" to "$5$" (the 4th image in the 2nd row of Figure~\ref{visual_s}), and numbers "66" to "00" (the 8th image in the 2nd row of Figure~\ref{visual_s}). In comparison, the other adversarial methods also appear to generate adversarial examples in a same sense but not as obvious as MAT. This may partly explain why the adversarial examples would be difficult to be recognized by many traditional deep neural networks. Furthermore, generating the most challenging examples (being similar to or even changed to other categories), MAT indeed leads to the most serious adversarial attacks that deep neural networks hardly recognize. Finally, it is observed that the major changed areas occur on the number strokes, which can be considered as the ``manifold" of these digit images. This implies that our MAT training method could truly generate adversarial attacks in the manifold.

Last, we also show in Figure~\ref{visual_c} adversarial examples on CIFAR-10 generated from various methods. Unfortunately, very different from MNIST and SVHN, it appears that the resulting examples give no obvious clues on how the images would be changed. This might be explained by the fact that general objects are sufficiently more complicated than digits. We would leave the topic of generating explainable adversarial examples in the future.

\section{Conclusion}
We present the Manifold Adversarial Training (MAT), a novel method to smooth the distributional manifold in the latent space. Compared with other adversarial training methods, our proposed MAT learns adversarially a robust feature representation in the latent space, making the latent space both informative and discriminative. Specifically, we first represent the latent features with Gaussian mixtures. We then define the smoothness of distributional manifold based on the $KL$-divergence between Gaussian mixtures for the original and adversarial examples.  We implemented MAT on MNIST, CIFAR-10 and SVHN in both supervised and unsupervised tasks. The results showed that our proposed model is much better than those state-of-the-art methods.

\bibliographystyle{IEEEtran}
\bibliography{reference}

\end{document}